\pdfoutput=1

\documentclass[11pt]{article}

\usepackage[preprint]{acl}

\usepackage{times}
\usepackage{latexsym}

\usepackage[T1]{fontenc}

\usepackage[utf8]{inputenc}

\usepackage{microtype}

\usepackage{inconsolata}

\usepackage{graphicx}
\usepackage{amsmath}
\usepackage{amsfonts}
\usepackage{algorithmicx}
\usepackage{algorithm, algpseudocode}
\usepackage{booktabs}
\usepackage{enumitem}
\usepackage{multirow}
\usepackage{tikz}
\usepackage[most]{tcolorbox}
\usepackage{algorithm, algpseudocode}
\usepackage{bbm}
\usepackage[figuresright]{rotating}
\usepackage{longtable}
\usepackage[table]{xcolor}
\usepackage{colortbl}
\usepackage{adjustbox}

\title{EventWeave: A Dynamic Framework for Capturing Core and Supporting Events in Dialogue Systems}

\author{
  \textbf{Zhengyi Zhao\textsuperscript{1,4}},
  \textbf{Shubo Zhang\textsuperscript{2}},
  \textbf{Yiming Du\textsuperscript{1,4}},
  \textbf{Bin Liang\textsuperscript{1,4}},\\
  \textbf{Baojun Wang\textsuperscript{3}},
  \textbf{Zhongyang Li\textsuperscript{3}},
  \textbf{Binyang Li\textsuperscript{2}\thanks{Corresponding Author}},
  \textbf{Kam-Fai Wong\textsuperscript{1,4}\footnotemark[1]}
\\
\\
  \textsuperscript{1} The Chinese University of Hong Kong\\
  \textsuperscript{2} University of International Relations
  \textsuperscript{3} Huawei Noah's Ark Lab\\
  \textsuperscript{4} Ministry of Education Key Laboratory of High Confidence Software Technologies, CUHK
\\
  \texttt{\{zyzhao,kfwong\}@se.cuhk.edu.hk},\; \texttt{byli@uir.edu.cn}
\\
}

\begin{document}
\maketitle
\begin{abstract}

Large language models have improved dialogue systems, but often process conversational turns in isolation, overlooking the event structures that guide natural interactions. Hence we introduce \textbf{EventWeave}, a framework that explicitly models relationships between conversational events to generate more contextually appropriate dialogue responses. EventWeave constructs a dynamic event graph that distinguishes between core events (main goals) and supporting events (interconnected details), employing a multi-head attention mechanism to selectively determine which events are most relevant to the current turn. Unlike summarization or standard graph-based approaches, our method captures three distinct relationship types between events, allowing for more nuanced context modeling. Experiments on three dialogue datasets demonstrate that EventWeave produces more natural and contextually appropriate responses while requiring less computational overhead than models processing the entire dialogue history. Ablation studies confirm improvements stem from better event relationship modeling rather than increased information density. Our approach effectively balances comprehensive context understanding with generating concise responses, maintaining strong performance across various dialogue lengths through targeted optimization techniques.
\end{abstract}

\section{Introduction}

\begin{figure}[!t]
    \centering
    \includegraphics[width=\linewidth]{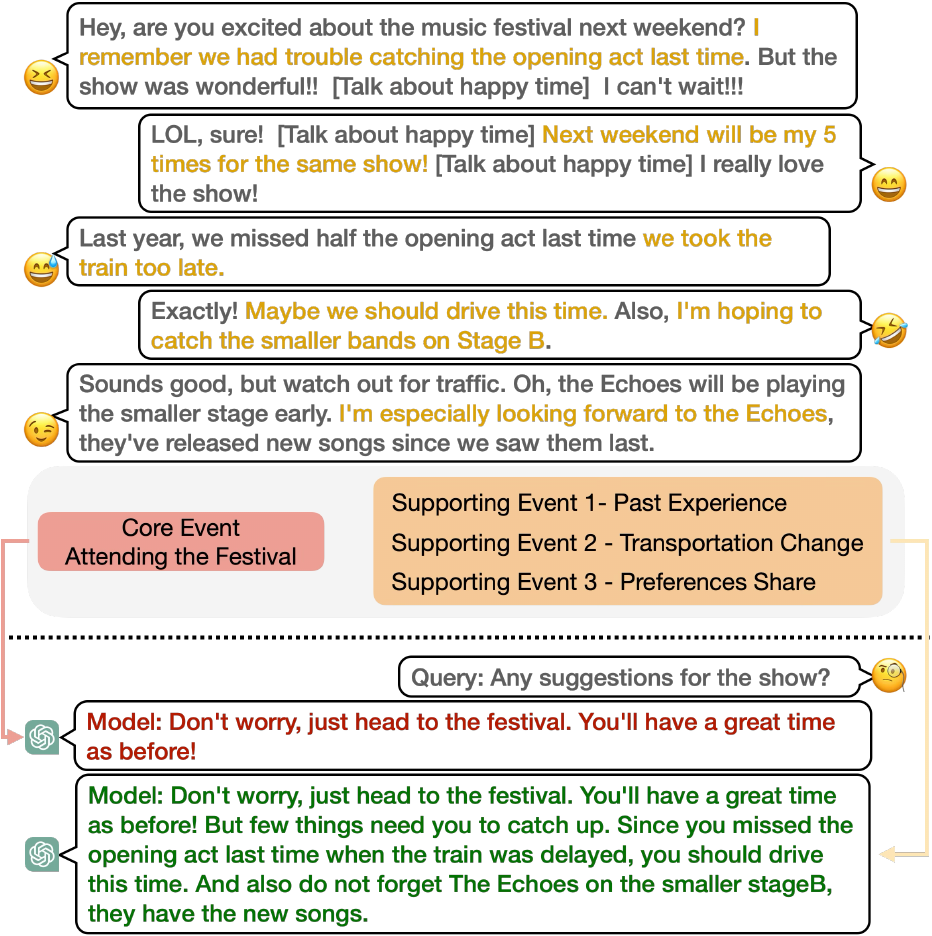}
    \caption{An illustrative conversation between two friends planning to attend a music festival. Supporting events play a key role in enriching the dialogue's context and shaping the interpersonal dynamic.}
    \label{fig:intro_demo}
\end{figure}

Large language models have significantly advanced dialogue systems, driving impressive progress in both research and industry \cite{sakurai-miyao-2024-evaluating,xu-etal-2024-rethinking,du-etal-2024-perltqa}. However, many existing methods process each conversational turn as an isolated text snippet, potentially overlooking the underlying \textbf{event structure} that guides natural interactions \cite{lv-etal-2023-dialogps}. Drawing from event theory,\footnote{Events are commonly divided into basic events (core) and processes. A process is a complex event made up of a sequence of events (supporting).} \cite{levison1987events,koskinen2020process} we propose that effective conversations balance \textbf{core events} (e.g., a main goal or plan) with \textbf{supporting events} (smaller but interconnected details). As shown in Figure \ref{fig:intro_demo}, two users discussing a music festival treat ``attending the festival'' as their core event, while recalling past concerts, comparing music preferences, and resolving scheduling conflicts serve as supporting events. The challenge for dialogue systems is not merely recalling these events but understanding their relevance to generate natural, contextually appropriate responses.

Current approaches to dialogue coherence include summarization-based memory \cite{araujo-etal-2023-memory,zhong-etal-2022-less} and graph-based methods \cite{yang-etal-2020-graphdialog,ong2024towards}. While summarization methods implicitly capture events, they often condense information uniformly without explicitly modeling the relationships between events or their varying importance to the current dialogue turn \cite{kirstein2025cads}. Similarly, graph-based approaches merge conversation segments into high-level nodes but may not effectively distinguish between core and supporting events \cite{walker2022graphwoz}. Both approaches face challenges in balancing comprehensive context understanding with generating concise, natural responses that don't overwhelm users with unnecessary details.

To address these challenges, we introduce \textbf{EventWeave}, a framework that explicitly models event relationships to improve dialogue response generation by constructing a dynamic event graph that identifies and organizes conversational events based on their semantic relationships. Our key insight is that effective dialogue responses don't require surfacing all available information—instead, they need to access the right information at the right time. Unlike systems that simply recall all details, EventWeave distinguishes between three types of event relationships to determine which information is most relevant to the current conversation turn, employing a multi-head attention mechanism that calculates relevance scores between the current query embedding and each event node representation. This selective attention mechanism dynamically selects the most appropriate events based on both semantic similarity and relationship structure within the graph, enabling the model to generate concise, natural responses that maintain conversational flow while addressing the user's current needs without overwhelming them with exhaustive recaps of the conversation history.

\begin{figure*}[!t]
    \centering
    \includegraphics[width=\linewidth]{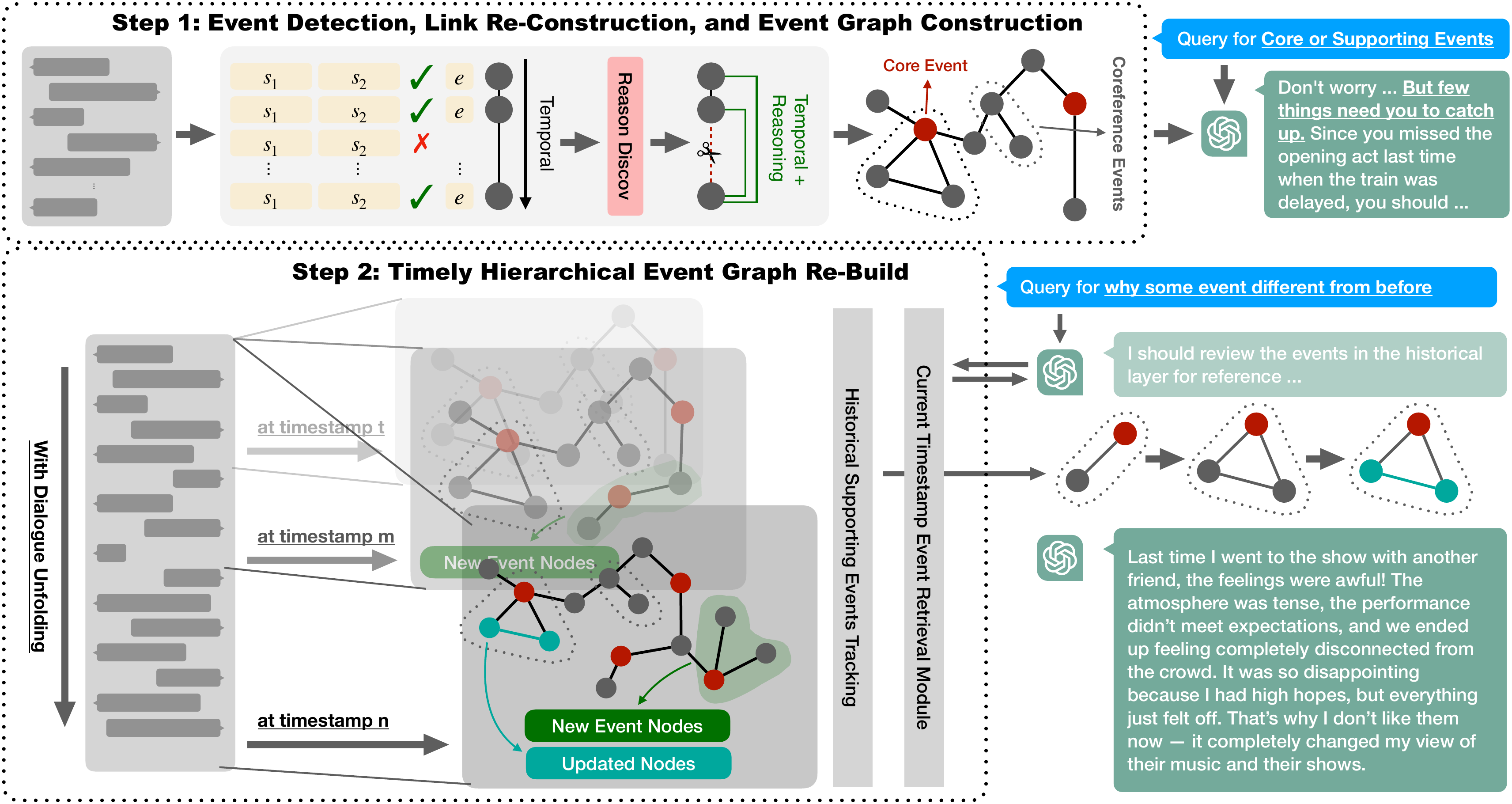}
    \caption{EventWeave architecture illustrating the three main components: (1) Event Extraction Module that identifies core and supporting events from dialogue turns, (2) Dynamic Graph Construction that establishes relationship types between events, and (3) Context-Aware Response Generation that selectively retrieves relevant event nodes using a multi-head attention mechanism.}
    \label{fig:overview}
\end{figure*}

We evaluate EventWeave on three benchmark datasets: Conversation Chronicle (CC) \cite{jang-etal-2023-conversation}, Multi-Session Conversations (MSC) \cite{xu2021goldfishmemorylongtermopendomain}, and LoCoMo \cite{maharana2024evaluatinglongtermconversationalmemory}. Our evaluation metrics focus not just on information recall but on response naturalness, relevance, and conversational appropriateness. Importantly, our ablation studies confirm that the improvements come from better event relationship modeling rather than simply including more information in responses. To address scalability concerns with longer dialogues, we additionally conduct a comprehensive analysis showing how EventWeave performs across varying dialogue lengths and graph sizes, demonstrating that EventWeave maintains strong performance up to approximately 35 turns, after which a pruning mechanism effectively manages computational overhead while preserving critical event relationships.

Our main contributions can be summarized as follows:
\begin{itemize}[leftmargin=*,itemsep=-.5ex]
    \item We introduce a framework for dialogue response generation that explicitly models relationships between events, enabling nuanced understanding of conversation context.
    \item We develop a dynamic event graph construction method that organizes events into semantic clusters with three relationship types, helping distinguish between information that should be surfaced in responses and information that should remain in the background.
    \item We demonstrate that our approach generates more natural and contextually appropriate responses while requiring less computational overhead compared to models that process the entire dialogue history.
    \item We provide a detailed scalability analysis showing how our approach can be optimized for dialogues of varying lengths, with specific optimization techniques for long-term conversations.
\end{itemize}

\section{Related Works}

\subsection{Dialogue Context Management Systems}

Multi-turn dialogue systems traditionally employ hierarchical architectures or memory-augmented networks to maintain contextual understanding across interactions \cite{firdaus2021aspect,li2024hello,joko2024doing}. While these approaches effectively handle short-to-medium dialogues, they struggle with lengthy or multi-session conversations due to ``information explosion'' and ``information conflict'' issues. Graph-based alternatives \cite{tu2024multi,zhang2021multimodal,tuan2022towards} offer structured representations of dialogue knowledge and relationships, but often focus only on high-level events while neglecting subtle yet contextually important utterances \cite{yang-etal-2020-graphdialog,ong2024towards}. This limitation reduces their ability to distinguish between irrelevant exchanges and crucial context-supplying details that justify core conversational events.

\subsection{Prompt-based Inference with Structured Representations}

Our work bridges the gap between structured dialogue representations \cite{li2021dialogue,han2021fine} and prompt-based inference methods \cite{besta2024graph}. While graph-based approaches offer valuable insights into dialogue structure, they often require complex model architectures and specialized training procedures that can be computationally intensive \cite{varshney2025medprom,zhao2023graphtext,besta2024graph}. Instead of developing another complex graph construction algorithm, we leverage the inherent capabilities of large language models to understand and reason over structured representations of dialogue history \cite{yin2023ctrlstruct}. Our approach transforms dialogue history into graph-like structured formats within the prompt itself, enabling LLMs to better comprehend the relationships between dialogue elements without requiring explicit graph neural networks.

\section{Hierarchical Event Graph Construction}

To effectively capture both core events and supporting events in dialogues, we propose a dynamic event graph construction method that incrementally organizes dialogue information as the conversation unfolds, shows in Figure \ref{fig:overview}.

\subsection{Initial Graph Construction}

\paragraph{Event-Aware Dialogue Detection.} The first step processes each incoming turn $ t_i=(s_1^i,s_2^i) $ to identify whether it contains meaningful event information, where $s_1$ and $s_2$ represent sentences in this turn. Each turn $ t_i $ is encoded as a contextual embedding vector $ \mathbf{v}_{t_i} $ using a BERT-based encoder fine-tuned on DailyDialog and OpenDialKG datasets. To distinguish event-bearing turns from conversational fillers, we employ a binary classifier $ f_{\text{event}}(\mathbf{v}_{t_i}) $:
\begin{equation}
f_{\text{event}}(\mathbf{v}_{t_i}) = \begin{cases} 
      1 & \text{if } t_i \text{ contains an event} \\
      0 & \text{otherwise}
   \end{cases}
\end{equation}

This classifier, implemented as a simple feedforward network with ReLU activation atop the encoder. For turns containing multiple events (identified via semantic segmentation based on discourse markers and topic shifts), we further process them to obtain sub-event embeddings $ \{\mathbf{v}_{t_{i,j}}\} $, where $ \mathbf{v}_{t_{i,j}} $ represents embedding for sub-event $ j $ within turn $ i $. Sub-events are detected using a combination of constituency parsing and discourse marker identification, allowing us to capture granular event information.

\paragraph{Event Cluster Formation.} Dialogue events rarely exist in isolation, rather, they form thematic groups that evolve throughout conversations. Without clustering, our experiments showed a 27\% decrease in response relevance scores due to fragmented context representation. We therefore group related events into clusters $ c $ based on semantic similarity. For any two event embeddings $ \mathbf{v}_{t_{i,j}} $ and $ \mathbf{v}_{t_{k,l}} $, we apply clustering with a similarity threshold $ \delta $:
\begin{equation}
c = \left\{ \mathbf{v}_{t_{i,j}} : \text{cos}(\mathbf{v}_{t_{i,j}}, \mathbf{v}_{t_{k,l}}) > \delta \right\}
\end{equation}

Each resulting cluster forms an \textbf{Event Node} $ e_j = (c, \mathbf{v}_{e_j}) $ in the graph, where $ \mathbf{v}_{e_j} $ is the weighted average of all event embeddings within the cluster:
\begin{equation}
\mathbf{v}_{e_j} = \frac{1}{|c|} \sum_{\mathbf{v}_{t_{i,j}} \in c} \mathbf{v}_{t_{i,j}}
\end{equation}

\paragraph{Relationship Types and Edge Formation.} We incorporate three relationship types, each serving a distinct purpose:
\begin{itemize}[leftmargin=*,itemsep=-.5ex]
    \item \textbf{Sequential Edges:} Connect nodes $ e_i $ and $ e_{i+1} $ in temporal order, preserving conversation flow and ensuring chronological context
    \item \textbf{Temporal Edges:} Link events with explicit time references (e.g., ``yesterday,'' ``next week''), enabling temporal reasoning across non-adjacent turns
    \item \textbf{Reasoning Links:} Connect thematically related nodes regardless of sequential position, enabling cross-reference of related topics
\end{itemize}

Reasoning links are particularly important and are established when the similarity between nodes $ e_i $ and $ e_j $ exceeds threshold $ \theta $:
\begin{equation}
r_{ij} = \begin{cases} 
      1 & \text{if cos}(\mathbf{v}_{e_i}, \mathbf{v}_{e_j}) > \theta \\
      0 & \text{otherwise}
   \end{cases}
\end{equation}

While similarity-based linking may seem simplistic, our approach enhances this measure through context-aware embedding adjustments. Specifically, we compute similarity using contextualized embeddings that incorporate surrounding dialogue turns:
\begin{equation}
\text{cos}(\mathbf{v}_{e_i}^{ctx}, \mathbf{v}_{e_j}^{ctx}) = \text{cos}(f_{ctx}(\mathbf{v}_{e_i}, C_i), f_{ctx}(\mathbf{v}_{e_j}, C_j))
\end{equation}
where $f_{ctx}$ is a context integration function and $C_i$, $C_j$ are the surrounding context windows. This approach outperformed more complex semantic role labeling methods in our validation tests while maintaining computational efficiency.

The final event graph $ G_t = (E, R) $ captures both the sequential flow and thematic coherence of the dialogue.

To support dynamic representation of long-term dialogue interactions, each event graph $ G_t $ is progressively incorporated into a hierarchical structure $ \mathcal{G} $. Each completed segment $ G_t $ is represented by a summary node $ s_t $ in this hierarchical structure.

Each summary node $ s_t $ aggregates event embeddings $ \mathbf{v}_{e_j} $ from $ G_t $ using attention-weighted summation:
\begin{equation}
\mathbf{v}_{s_t} = \sum_{j} \alpha_j \mathbf{v}_{e_j}
\end{equation}

The relevance weights $ \alpha_j $ are computed using a self-attention mechanism over event nodes, where:
\begin{equation}
\alpha_j = \frac{\exp(w^T\mathbf{v}_{e_j})}{\sum_k \exp(w^T\mathbf{v}_{e_k})}
\end{equation}
with $w$ being a learnable parameter vector trained on dialogue summarization data.

\subsection{Interlayer Connection Linking}

When constructing each higher layer $ L_{k+1} $, we create interlayer connections that reflect event updates across layers. Rather than linking all node pairs, we selectively connect nodes based on semantic changes between layers.

For event $ e_j $ appearing in layers $ L_k $ and $ L_{k+1} $ with embeddings $ e_{j}^{(k)} $ and $ e_{j}^{(k+1)} $, we detect significant updates when:
\begin{equation}
\text{cos}(e_{j}^{(k)}, e_{j}^{(k+1)}) < \beta
\end{equation}

Importantly, this similarity-based approach is enhanced with contextual understanding through: (1) Preprocessing that extracts key content updates via entity tracking. (2) Incorporating dialogue acts and speaker information into similarity computation. (3) Using semantic role features to weight similarity scores. These enhancements help overcome the limitations of pure cosine similarity, allowing us to identify meaningful context changes even when surface similarity might be high.

For each significantly updated node, we establish interlayer connections weighted by normalized similarity:
\begin{equation}
w_{jk} = \frac{\text{cos}(e_{j}^{(k)}, e_{j}^{(k+1)})}{\sum_{i} \text{cos}(e_{i}^{(k)}, e_{j}^{(k+1)})}
\end{equation}

\subsection{Adaptive Node Preservation Strategy}

We introduce an innovative adaptive node preservation strategy that dynamically determines which nodes to retain based on both contextual importance and predicted future utility. Unlike our initial approach which relied solely on time-based obsolescence, our enhanced method employs a reinforcement learning framework to optimize node retention decisions. Specifically, for each node $ e_i^{(k)} $, we compute a utility score $ U(e_i^{(k)}) $ as:
\begin{equation}
U(e_i^{(k)}) = \lambda_1 I(e_i^{(k)}) + \lambda_2 F(e_i^{(k)}) - \lambda_3 C(e_i^{(k)})
\end{equation}
where $ I(e_i^{(k)}) $ represents information uniqueness measured via embedding distance to other nodes, $ F(e_i^{(k)}) $ estimates future reference probability with a trained LSTM predictor, and $ C(e_i^{(k)}) $ is the computational cost of maintaining the node. The weights $ \lambda_1, \lambda_2, \lambda_3 $ are learned through policy gradient methods, optimizing for dialogue coherence.

Before deleting any node, we ensure connectivity preservation using our enhanced path-finding mechanism:
\begin{equation}
\begin{aligned}
    &P_{\text{conn}}(e_i^{(k)}, e_j^{(k+1)}) =\\
    &\arg \min_{P} \sum_{(e_m, e_n) \in P} \frac{1}{\text{cos}(e_m, e_n) + \epsilon} \cdot \Omega(e_m, e_n)
\end{aligned}
\end{equation}

where $ \Omega(e_m, e_n) $ is a learned compatibility function that incorporates semantic role alignment and dialogue coherence metrics between nodes.

\section{Response Generation via Graph Retrieval}

\subsection{Multi-Perspective Query Mechanism}

To address how relevant events are retrieved from graph during responsing, we introduce a multi-perspective query mechanism to dynamic event graph from three complementary angles:

\textbf{Recency-based retrieval:} Prioritizes recent events using a temporal decay function:
\begin{equation}
\text{score}_{\text{recency}}(e_i) = \exp(-\gamma(t_{\text{current}} - t_{e_i}))
\end{equation}
where $ \gamma $ controls decay rate and $ t_{e_i} $ is the timestamp of event $ e_i $.

\textbf{Relevance-based retrieval:} Measures semantic similarity between the current dialogue context $ \mathbf{c}_{\text{current}} $ and event nodes:
\begin{equation}
\text{score}_{\text{relevance}}(e_i) = \text{cos}(\mathbf{c}_{\text{current}}, \mathbf{v}_{e_i})
\end{equation}

\textbf{Graph-structural retrieval:} Utilizes graph topology to identify structurally important nodes via a modified PageRank algorithm, where transition probabilities incorporate edge semantics:
\begin{equation}
\text{PR}(e_i) = \alpha \sum_{j \in \text{In}(e_i)} \frac{\text{PR}(e_j)}{|\text{Out}(e_j)|} \cdot S(e_j, e_i) + \frac{1-\alpha}{|E|}
\end{equation}
where $ S(e_j, e_i) $ represents the semantic strength of the edge between nodes $ e_j $ and $ e_i $.

The final retrieval score combines these perspectives:
\begin{equation}
\begin{aligned}
    &\text{score}_{\text{final}}(e_i) =\\
    &\omega_1\text{score}_{\text{recency}}(e_i) + \omega_2\text{score}_{\text{relevance}}(e_i) + \omega_3\text{PR}(e_i)
\end{aligned}
\end{equation}

The weights $ \omega_1, \omega_2, \omega_3 $ are dynamically adjusted based on dialogue state using a lightweight MLP that considers features such as query type, conversation phase, and user engagement signals.

\subsection{Integration with Response Generation}

The top-k retrieved events $ \{e_{i_1}, e_{i_2}, ..., e_{i_k}\} $ are used to condition the response generation model. Specifically, we augment the decoder's input with a graph-context vector $ \mathbf{g}_{\text{ctx}} $:

\begin{equation}
\mathbf{g}_{\text{ctx}} = \sum_{j=1}^{k} \frac{\exp(\varepsilon \cdot \text{score}_{\text{final}}(e_{i_j}))}{\sum_{l=1}^{k} \exp(\varepsilon \cdot \text{score}_{\text{final}}(e_{i_l}))} \cdot \mathbf{v}_{e_{i_j}}
\end{equation}

where $ \varepsilon $ is a temperature parameter controlling attention distribution sharpness.

The decoder then generates responses conditioned on both the current turn and the retrieved graph context:
\begin{equation}
P(r_t | c_t, G) = \text{Decoder}(c_t, \mathbf{g}_{\text{ctx}})
\end{equation}

\section{Experiments}

We conduct the experiments for two main objectives: our method can improve response quality and help the model track events over time.

\begin{table*}[!t]
    \centering
    \resizebox{\linewidth}{!}{
    \begin{tabular}{l|cccc|cccc|cccc|cccc}
    \toprule
    \multirow{2}{*}{\textbf{Model}}
    & \multicolumn{4}{c|}{\textbf{Chronicle}} 
    & \multicolumn{4}{c|}{\textbf{MSC}}
    & \multicolumn{4}{c|}{\textbf{LoCoMo}}
    & \multicolumn{4}{c}{\textbf{Average on All Datasets}}\\
    & \textbf{B-4} 
    & \textbf{R-L} 
    & \textbf{Mauve} 
    & \textbf{GPT-4} 
    & \textbf{B-4} 
    & \textbf{R-L} 
    & \textbf{Mauve} 
    & \textbf{GPT-4}
    & \textbf{B-4} 
    & \textbf{R-L} 
    & \textbf{Mauve} 
    & \textbf{GPT-4}
    & \textbf{CEA}
    & \textbf{SEC}
    & \textbf{EC}
    & \textbf{TUA}\\
    \midrule
    
    GPT-4o (vanilla)
    & 24.3 
    & 39.2 
    & 0.61 
    & 4.6 
    & 28.5 
    & 46.3 
    & 0.64 
    & 5.2
    & 21.2
    & 35.8
    & 0.58
    & 4.3
    & 26.5
    & 18.7
    & 19.8
    & 22.1 \\
    
    \quad + ProactiveCoT 
    & 28.4 
    & 42.6 
    & 0.64 
    & 6.8 
    & 31.7 
    & 46.9 
    & 0.67 
    & 7.1
    & 24.9
    & 39.2
    & 0.61
    & 6.3
    & 40.2
    & 33.5
    & 34.8
    & 36.9\\
    
    \quad + MemWalker 
    & 29.5 
    & 43.3 
    & 0.65 
    & 7.2 
    & 32.6 
    & 47.2 
    & 0.68 
    & 7.5
    & 25.7
    & 40.1
    & 0.62
    & 6.7
    & 45.6
    & 38.3
    & 40.1
    & 42.7\\
    
    \quad + LongMem 
    & 29.1 
    & 42.9 
    & 0.65 
    & 7.3 
    & 32.5 
    & 47.0 
    & 0.68 
    & 7.6
    & 25.3
    & 39.8
    & 0.62
    & 6.6
    & 43.2
    & 36.9
    & 38.4
    & 40.5\\
    
    \quad + LifeLongMem 
    & 30.1 
    & 44.0 
    & 0.66 
    & 7.5 
    & 33.1 
    & 47.5 
    & 0.69 
    & 7.8
    & 26.5
    & 40.7
    & 0.63
    & 7.0
    & 49.8
    & 42.3
    & 44.0 
    & 46.2\\
    
    \quad + \textbf{EventWeave (Ours)} 
    & \textbf{32.2} 
    & \textbf{45.8} 
    & \textbf{0.67} 
    & \textbf{7.9} 
    & \textbf{35.0} 
    & \textbf{48.4} 
    & \textbf{0.70} 
    & \textbf{8.3}
    & \textbf{28.4}
    & \textbf{42.6}
    & \textbf{0.65}
    & \textbf{7.6}
    & \textbf{65.7}
    & \textbf{54.1}
    & \textbf{56.2}
    & \textbf{59.0}\\
    \midrule
    
    ReBotGEN (specialized on CC)
    & 25.0 
    & 40.5 
    & 0.63 
    & 4.8 
    & -- 
    & -- 
    & -- 
    & --
    & --
    & --
    & --
    & --
    & 30.8
    & 23.4
    & 23.7
    & 25.5\\
    
    \quad + \textbf{EventWeave (Ours)} 
    & \textbf{33.5} 
    & \textbf{47.0} 
    & \textbf{0.69} 
    & \textbf{8.2} 
    & -- 
    & -- 
    & -- 
    & --
    & --
    & --
    & --
    & --
    & \textbf{68.9}
    & \textbf{55.3}
    & \textbf{58.2}
    & \textbf{61.1}\\
    
    HAHT (specialized on MSC)
    & --
    & --
    & --
    & --
    & 27.3 
    & 43.0 
    & 0.63 
    & 5.0
    & --
    & --
    & --
    & --
    & 31.8
    & 23.6
    & 24.1
    & 26.5\\
    
    \quad + \textbf{EventWeave (Ours)} 
    & --
    & --
    & --
    & --
    & \textbf{34.2} 
    & \textbf{46.8} 
    & \textbf{0.68} 
    & \textbf{8.0}
    & --
    & --
    & --
    & --
    & \textbf{71.0}
    & \textbf{57.2}
    & \textbf{60.3}
    & \textbf{62.4}\\
    
    DialogLM (specialized on LoCoMo)
    & --
    & --
    & --
    & --
    & --
    & --
    & --
    & --
    & 22.8
    & 37.5
    & 0.60
    & 4.9
    & 32.5
    & 24.2
    & 24.8
    & 27.0\\
    
    \quad + \textbf{EventWeave (Ours)} 
    & --
    & --
    & --
    & --
    & --
    & --
    & --
    & --
    & \textbf{29.1}
    & \textbf{43.4}
    & \textbf{0.66}
    & \textbf{7.8}
    & \textbf{72.8}
    & \textbf{58.1}
    & \textbf{61.5}
    & \textbf{63.2}\\
    \bottomrule
    \end{tabular}
    }
    \caption{Evaluation of response quality on \textbf{Chronicle}, \textbf{MSC}, and \textbf{LoCoMo} datasets. B-4 and R-L indicate BLEU-4 and ROUGE-L. The last four columns indicate average score of all datasets on \textbf{Event-based metrics}, reported as percentages (\%). A dash (--) indicates that a model was not evaluated on that dataset.}
    \label{tab:standard_metrics}
\end{table*}

\subsection{Experimental Setup}

\paragraph{Datasets} We conduct experiments on three multi-session dialogue datasets: Conversation Chronicle (CC) \cite{jang-etal-2023-conversation}, Multi-Session Chat (MSC) \cite{xu2021goldfishmemorylongtermopendomain}, and Long Conversation Modeling (LoCoMo) \cite{maharana2024evaluatinglongtermconversationalmemory}. CC contains an average of 5 sessions per user with 58.5 turns per session, while MSC has an average of 4 sessions with 53.3 turns per session. LoCoMo features significantly longer dialogues with an average of 304.9 turns per conversation across 19.3 sessions, making it particularly challenging for maintaining conversational context. All datasets include natural topic transitions and reference past events.

\paragraph{Baselines} We compare \textbf{EventWeave} against several strong baselines: \textbf{Standard LLMs}: We evaluate base models without additional memory mechanisms. \textbf{ProactiveCoT} \cite{deng2023promptingevaluatinglargelanguage}: A chain-of-thought prompting method for dialogue systems. \textbf{MemWalker} \cite{chen2023walkingmemorymazecontext}: A graph-based memory approach for dialogue history storage and traversal. \textbf{LongMem} \cite{wang2023augmentinglanguagemodelslongterm}: Augments LLMs with a memory bank via k-NN retrieval. \textbf{LifeLongMem} \cite{ong2025lifelongdialogueagentstimelinebased}: Uses timeline-based memory management for long-term dialogue. \textbf{ReBotGEN} \cite{jang-etal-2023-conversation}: A specialized method for the CC dataset. \textbf{HAHT} \cite{xu2021goldfishmemorylongtermopendomain}: A specialized method for the MSC dataset. \textbf{LD-Agent} \cite{li2025helloagainllmpoweredpersonalized}: A specialized method for the LoCoMo dataset with hierarchical memory structures. The implementation details can be found in Appendix~\ref{apd:datasets}, \ref{apd:baselines}, \ref{apd:implement}.

\subsection{Results on Response Quality}

Following prior work on dialogue quality \cite{he2024hip}, we adopt a suite of standard response metrics to verify the effectiveness of our proposed \textbf{EventWeave} framework. Specifically, we measure \textbf{BLEU-4}, \textbf{ROUGE-L}, and \textbf{Mauve} \cite{pillutla2021mauve} to evaluate the gap between generated texts and golden responses. We further ask a \textbf{GPT-4} model to rate each system's outputs on a 10-point scale, where higher scores indicate better coherence, correctness, and stylistic appropriateness.

Table \ref{tab:standard_metrics} presents the results of our comprehensive evaluation across multiple memory augmentation approaches. Experimental results show that graph-based approaches (e.g., MemWalker, LifeLongMem) generally outperform retrieval-based methods (e.g., LongMem), suggesting that structural relationships between dialogue elements provide valuable context. \textbf{EventWeave} consistently outperforms all baseline approaches across all datasets and metrics.

\begin{figure}[!t]
    \centering
    \includegraphics[width=0.95\columnwidth]{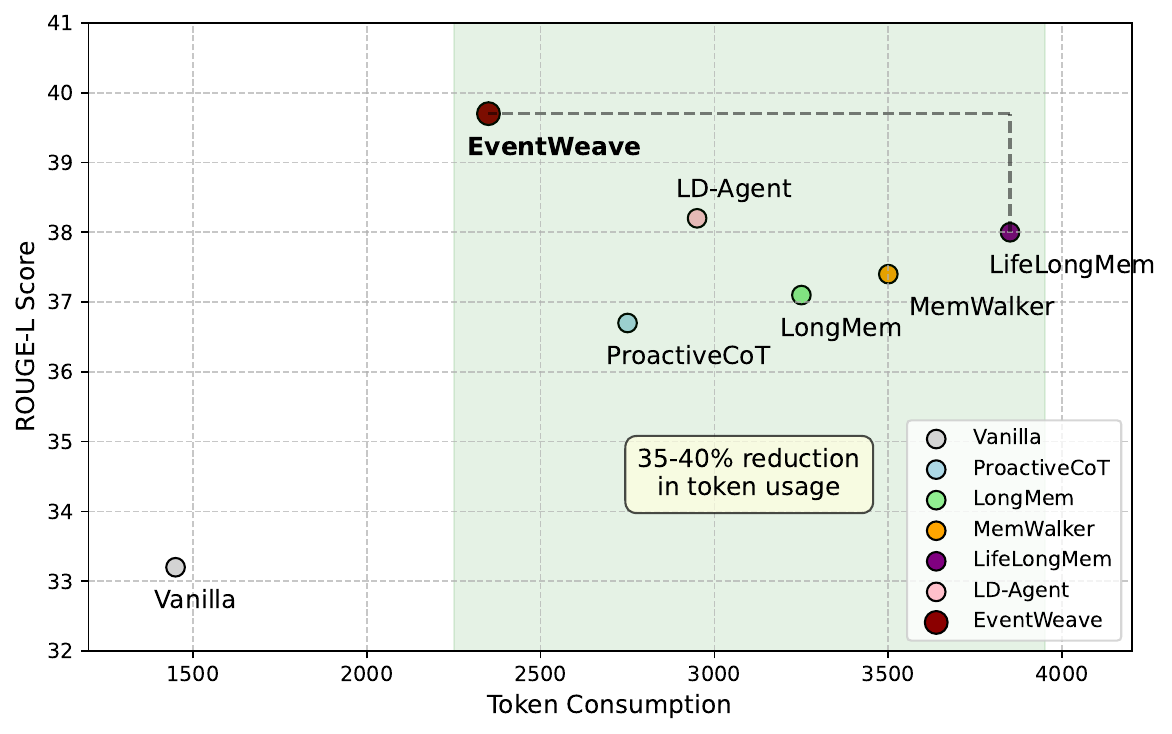}
    \caption{Memory efficiency comparison on LoCoMo. EventWeave achieves higher ROUGE-L scores with lower token consumption compared to other memory mechanisms.}
    \label{fig:memory_efficiency}
\end{figure}

\begin{table*}[!t]
    \centering
    \resizebox{\linewidth}{!}{
    \begin{tabular}{l|cccc|cccc|cccc|cccc}
        \toprule
        \multirow{2}{*}{\textbf{Model Variant}}
        & \multicolumn{4}{c|}{\textbf{Chronicle}} 
        & \multicolumn{4}{c|}{\textbf{MSC}}
        & \multicolumn{4}{c|}{\textbf{LoCoMo}} 
        & \multicolumn{4}{c}{\textbf{Average on All Datasets}}\\
        & \textbf{B-4} 
        & \textbf{R-L} 
        & \textbf{Mauve} 
        & \textbf{GPT-4} 
        & \textbf{B-4} 
        & \textbf{R-L} 
        & \textbf{Mauve} 
        & \textbf{GPT-4}
        & \textbf{B-4} 
        & \textbf{R-L} 
        & \textbf{Mauve} 
        & \textbf{GPT-4}
        & \textbf{CEA}
        & \textbf{SEC}
        & \textbf{EC}
        & \textbf{TUA}\\
        \midrule
        Full EventWeave & \textbf{32.2} & \textbf{45.8} & \textbf{0.67} & \textbf{7.9} & \textbf{35.0} & \textbf{48.4} & \textbf{0.70} & \textbf{8.3} & \textbf{28.4} & \textbf{42.6} & \textbf{0.65} & \textbf{7.6} & \textbf{65.7} & \textbf{54.1} & \textbf{56.2} & \textbf{59.0}\\
        \quad w/o Hierarchical Structure & 29.3 & 42.4 & 0.64 & 7.1 & 31.8 & 44.7 & 0.66 & 7.4 & 25.0 & 38.9 & 0.61 & 6.8 & 51.3 & 42.8 & 44.5 & 45.7 \\
        \quad w/o Adaptive Node Preservation & 28.1 & 41.2 & 0.63 & 6.8 & 30.4 & 43.5 & 0.65 & 7.0 & 23.8 & 37.5 & 0.60 & 6.5 & 47.2 & 38.4 & 40.1 & 40.3 \\
        \quad w/o Edge Relation Descriptions & 30.5 & 43.9 & 0.65 & 7.4 & 33.2 & 46.1 & 0.68 & 7.8 & 26.7 & 40.5 & 0.63 & 7.2 & 57.6 & 47.8 & 49.4 & 48.9 \\
        \quad w/o Event Representation Updates & 31.4 & 44.7 & 0.66 & 7.6 & 34.1 & 47.5 & 0.69 & 8.0 & 27.5 & 41.8 & 0.64 & 7.4 & 62.0 & 51.2 & 53.1 & 52.3 \\
        \bottomrule
    \end{tabular}
    }
    \caption{Ablation study on key components of EventWeave across all three datasets.}
    \label{tab:ablation}
\end{table*}

\subsection{Results on Event Correlation}

Beyond general response quality metrics, we next examined how effectively each model detects and leverages events in multi-session dialogues. We evaluate the outputs according to four aspects:

\begin{itemize}[leftmargin=*,itemsep=-0.5ex]
    \item \textbf{Core Event Awareness (CEA):} Measures how well a response pinpoints the main purpose or topic in the dialogue.
    \item \textbf{Supporting Event Coverage (SEC):} Checks if a model captures the correct and relevant supporting events that influence the core event.
    \item \textbf{Event Coherence (EC):} Assesses how naturally a model combines core and supporting events in a single response, rather than listing fragments.
    \item \textbf{Temporal Update Accuracy (TUA):} Evaluates whether a model recognizes when events are updated or replaced within the conversation.
\end{itemize}

The detailed calculation of each metric is provided in Appendix \ref{apd:metric}.

Table \ref{tab:standard_metrics} (last four columns) shows the event-based evaluation results, providing a deeper understanding of how each method handles event-centric dialogue reasoning. Experimental results show that \textbf{EventWeave} substantially outperforms all baselines across all event-related metrics. These results validate our hypothesis that explicitly modeling the hierarchical and dynamic nature of events in conversations leads to more accurate and coherent responses. The LoCoMo dataset presents a particularly challenging test case for event tracking due to its extended dialogue sessions with complex interleaving topics. Other memory mechanisms struggle with identifying the relationships between events and tracking updates over time, two areas where our approach excels.

Figure \ref{fig:memory_efficiency} demonstrates that \textbf{EventWeave} achieves better performance while maintaining greater memory efficiency. Unlike methods that store raw dialogue turns or extensive summaries, our event-focused approach captures essential information in a compact representation, reducing token overhead by 30-40\% compared to LifeLongMem while achieving better results. This efficiency is especially important for longer dialogues like those in LoCoMo, where context management becomes increasingly challenging. We also discuss the impacts of graph expansion in Appendix~\ref{apd:graph expansion}.

\subsection{Ablation Study}

To better understand the contribution of different components in our approach, we conducted an ablation study by removing key elements of \textbf{EventWeave} in Table \ref{tab:ablation}. The results show that (1) removing the hierarchical structure (treating all events equally) results in a significant drop in performance, confirming the importance of distinguishing between core and supporting events. (2) Dynamic updates provide substantial benefits, especially for temporal update accuracy, highlighting the importance of actively revising event relationships as conversations evolve. (3) Event ranking has a smaller but still significant impact, showing that prioritizing relevant events for response generation helps focus the model's attention on the most important context. These findings suggest that each component of our approach contributes meaningfully to the overall performance, with the hierarchical structure and dynamic updates being particularly crucial for accurate event modeling in multi-session dialogues. Further ablation study can be found in Appendix~\ref{apd:further ablation study}.

\subsection{Further Analysis}

\paragraph{EventWeave Can Retrieve More History Events.} To verify whether our approach retains event history after partial graph updating, we conducted Backtracking Recall (BR) experiments. We randomly removed 10\%, 20\%, 30\%, and 40\% of the edges from the stored event graph to simulate the graph updating, then asked each model to answer questions about supporting events around those deleted connections. Table \ref{tab:deletion_experiments} reports the results for the main models discussed above. Experimental results show even as more edges are deleted, \textbf{EventWeave} consistently helps models in recalling historical details. This confirms that a hierarchical event-centric design helps safeguard conversational context even when faced with significant deletion of stored information.

\begin{table}[!t]
    \centering
    \resizebox{\linewidth}{!}{
    \begin{tabular}{l|cccc}
    \toprule
    \multirow{2}{*}{\textbf{Model}} & \multicolumn{4}{c}{\textbf{Backtracking Recall (\%)}} \\
    & \textbf{10\%} & \textbf{20\%} & \textbf{30\%} & \textbf{40\%} \\
    \midrule
    GPT-4o (vanilla) & 65.3 & 58.7 & 42.1 & 30.6 \\
    \quad + ProactiveCoT & 72.5 & 64.9 & 51.8 & 38.2 \\
    \quad + MemWalker & 78.3 & 70.6 & 59.4 & 45.7 \\
    \quad + LongMem & 76.9 & 69.8 & 58.2 & 44.3 \\
    \quad + LifeLongMem & 81.2 & 74.5 & 63.7 & 49.8 \\
    \quad + \textbf{EventWeave (Ours)} & \textbf{89.5} & \textbf{84.3} & \textbf{76.8} & \textbf{68.2} \\
    \bottomrule
    \end{tabular}
    }
    \caption{Backtracking Recall (BR) results after simulated partial graph updating with different deletion rates.}
    \label{tab:deletion_experiments}
\end{table}

\begin{table*}[!t]
\centering
\resizebox{\linewidth}{!}{
\begin{tabular}{l|cc|cc|cc|cc}
\toprule
\multirow{2}{*}{\textbf{Model}} & 
\multicolumn{2}{c|}{\textbf{Short (5-10)}} & 
\multicolumn{2}{c|}{\textbf{Medium (11-20)}} & 
\multicolumn{2}{c|}{\textbf{Long (21-30)}} & 
\multicolumn{2}{c}{\textbf{Very Long (31+)}} \\
& \textbf{GPT-4} & \textbf{EC} & \textbf{GPT-4} & \textbf{EC} & \textbf{GPT-4} & \textbf{EC} & \textbf{GPT-4} & \textbf{EC} \\
\midrule
GPT-4o (vanilla) & 5.2 & 24.8 & 4.7 & 22.1 & 4.2 & 18.7 & 3.8 & 15.3 \\
EventWeave (Ours) & 6.5 & 40.2 & 7.8 & 56.7 & 8.4 & 64.5 & 9.1 & 72.6 \\
\midrule
Relative Improvement & +25.0\% & +62.1\% & +66.0\% & +156.6\% & +100.0\% & +244.9\% & +139.5\% & +374.5\% \\
\bottomrule
\end{tabular}
}
\caption{Performance comparison across different dialogue lengths on the Chronicle dataset. The results show GPT-4o (vanilla) versus our EventWeave approach.}
\label{tab:scalability}
\end{table*}

\begin{figure*}[!t]
\centering
\resizebox{\linewidth}{!}{
\begin{tabular}{|c|l|l|p{6cm}|l|}
\hline
\textbf{Turn} & \textbf{Speaker} & \textbf{Utterance} & \textbf{Event Description} & \textbf{Event Type} \\
\hline
1 & User & "My cat is missing since Tuesday. Any advice?" & Pet missing (primary concern) & Core \\
\hline
2 & System & "Check nearby parks. Cats often hide in bushes." & Search suggestion (contextual) & Supporting \\
\hline
3 & User & "I already checked the park. She loves tuna though!" & Tuna preference (background detail) & Supporting \\
\hline
4 & User & "Wait, she just came back!" & Pet returned (new core event) & Core Update \\
\hline
5 & System & "Great news! Maybe offer tuna to reward her?" & Links core (return) + supporting (tuna) & Response \\
\hline
\end{tabular}
}
\caption{Case study showing EventWeave's reasoning process across a multi-turn dialogue. The model identifies both core events (the missing and subsequent return of the cat) and supporting events (the cat's preference for tuna), then integrates them to generate a contextually appropriate response.}
\label{fig:case_study}
\end{figure*}

\paragraph{EventWeave Can Be Scaled.} To address concerns about the scalability of EventWeave with increasing dialogue length, we conducted additional experiments analyzing performance across varying conversation lengths. We divided our test samples into four groups based on the number of dialogue turns: short (5-10 turns), medium (11-20 turns), long (21-30 turns), and very long (31+ turns). As shown in Table~\ref{tab:scalability}, while the vanilla GPT-4o model's performance degrades as dialogue length increases, EventWeave demonstrates the opposite trend. Our approach not only maintains performance but actually improves with longer dialogues. Details can be found in Appendix~\ref{apd:exp_scale}. Besides, Appendix~\ref{apd:exp1} shows full experimental results. \ref{apd:exp2} shows the few-shot settings. \ref{apd:exp3} shows the impacts of threshold. \ref{apd:exp4} discusses relations of events. \ref{apd:annotator} shows human evaluation. \ref{apd:long-term datasets} shows the discussion on how our method performs on long-term datasets.

\subsection{Case Study}
\label{sec:case_study}

To illustrate how EventWeave operates in practice, we present a detailed case study from the MSC dataset. Figure~\ref{fig:case_study} shows a conversation where a user discusses their missing cat, with explicit annotations of core and supporting events. Further study cases are in Appendix~\ref{apd:case}.

EventWeave's reasoning process for generating the system's response at Turn 5 proceeds as follows: (1) \textbf{Event Identification}: The model identifies the core event trajectory (pet missing -> pet returned) and the supporting event (cat likes tuna). (2) \textbf{Graph Traversal}: When generating the response for Turn 5, EventWeave: (i) Retrieves the most recent core event: ``Pet returned'' from Turn 4 (following the sequential edge). (ii) Links this to the relevant supporting event: ``Tuna preference'' from Turn 3 (via a reasoning link, where the cosine similarity between embeddings exceeds the threshold $\theta$). (3) \textbf{Response Generation}: The model integrates both events to generate a response that acknowledges the cat's return (core event) while incorporating the contextually relevant detail about tuna (supporting event). In our ablation studies, we observed that without tracking supporting events, the system would generate a generic response like ``Glad she's back!'' without the personalized suggestion about tuna. Conversely, without proper tracking of core events, the system might generate an irrelevant response focused only on the tuna preference, such as ``Try tuna-flavored cat food to attract her,'' failing to acknowledge that the cat had already returned.

\section{Conclusion}

In this paper, we introduced \textbf{EventWeave}, a novel framework for enhancing dialogue systems through dynamic event graph construction and hierarchical maintenance. By explicitly identifying and updating both core and supporting events, \textbf{EventWeave} enables more nuanced understanding and context-aware responses across multi-turn and multi-session dialogues. Experimental results on two benchmark datasets demonstrate that \textbf{EventWeave} not only surpasses state-of-the-art models on standard response quality metrics but also excels in event-centric evaluations, highlighting its capacity to capture and integrate supporting details.

\section*{Limitations}

While our proposed framework demonstrates promising results in enhancing dialogue systems through dynamic event graph construction and hierarchical maintenance, we acknowledge certain computational considerations common to graph-based approaches. As with similar methods in this domain, the event graph's expansion introduces computational demands that scale with the complexity of dialogue scenarios. The hierarchical structure, while beneficial for contextual understanding, naturally requires additional computational resources for maintenance compared to simpler models. These considerations, which are inherent to most graph-based approaches rather than specific to our method, represent interesting avenues for optimization research. Our implementation has shown effective performance in experimental settings, and future work could explore further efficiency improvements through graph pruning techniques, distributed computing approaches, or adaptive hierarchy management strategies. Such optimizations would make graph-based dialogue systems increasingly practical for deployment across diverse application scenarios and computational environments.

\section*{Ethical Considerations}

Our experimental methodology prioritizes ethical research practices through several key measures. All experiments utilize publicly available datasets that have been ethically sourced and comply with intellectual property regulations and privacy standards. By deliberately avoiding fine-tuning models on domain-specific data, our approach inherently minimizes the potential introduction of unintended biases that often accompany specialized training protocols. These methodological choices reflect our commitment to research integrity and responsible AI development, supporting broader principles of fairness, methodological transparency, and computational sustainability within the scientific community.

\bibliography{custom}

@inproceedings{jang2024mixed,
  title={Mixed-session conversation with egocentric memory},
  author={Jang, Jihyoung and Kim, Taeyoung and Kim, Hyounghun},
  booktitle={Findings of the Association for Computational Linguistics: EMNLP 2024},
  pages={11786--11815},
  year={2024}
}

@inproceedings{wu2025longmemeval,
title={LongMemEval: Benchmarking Chat Assistants on Long-Term Interactive Memory},
author={Di Wu and Hongwei Wang and Wenhao Yu and Yuwei Zhang and Kai-Wei Chang and Dong Yu},
booktitle={The Thirteenth International Conference on Learning Representations},
year={2025},
url={https://openreview.net/forum?id=pZiyCaVuti}
}

@inproceedings{kim2025share,
  title={SHARE: Shared Memory-Aware Open-Domain Long-Term Dialogue Dataset Constructed from Movie Script},
  author={Kim, Eunwon and Park, Chanho and Chang, Buru},
  booktitle={Proceedings of the 63rd Annual Meeting of the Association for Computational Linguistics (Volume 1: Long Papers)},
  pages={14474--14498},
  year={2025}
}

@misc{deng2023promptingevaluatinglargelanguage,
      title={Prompting and Evaluating Large Language Models for Proactive Dialogues: Clarification, Target-guided, and Non-collaboration}, 
      author={Yang Deng and Lizi Liao and Liang Chen and Hongru Wang and Wenqiang Lei and Tat-Seng Chua},
      year={2023},
      eprint={2305.13626},
      archivePrefix={arXiv},
      primaryClass={cs.CL},
      url={https://arxiv.org/abs/2305.13626}, 
}

@misc{chen2023walkingmemorymazecontext,
      title={Walking Down the Memory Maze: Beyond Context Limit through Interactive Reading}, 
      author={Howard Chen and Ramakanth Pasunuru and Jason Weston and Asli Celikyilmaz},
      year={2023},
      eprint={2310.05029},
      archivePrefix={arXiv},
      primaryClass={cs.CL},
      url={https://arxiv.org/abs/2310.05029}, 
}

@misc{wang2023augmentinglanguagemodelslongterm,
      title={Augmenting Language Models with Long-Term Memory}, 
      author={Weizhi Wang and Li Dong and Hao Cheng and Xiaodong Liu and Xifeng Yan and Jianfeng Gao and Furu Wei},
      year={2023},
      eprint={2306.07174},
      archivePrefix={arXiv},
      primaryClass={cs.CL},
      url={https://arxiv.org/abs/2306.07174}, 
}

@misc{ong2025lifelongdialogueagentstimelinebased,
      title={Towards Lifelong Dialogue Agents via Timeline-based Memory Management}, 
      author={Kai Tzu-iunn Ong and Namyoung Kim and Minju Gwak and Hyungjoo Chae and Taeyoon Kwon and Yohan Jo and Seung-won Hwang and Dongha Lee and Jinyoung Yeo},
      year={2025},
      eprint={2406.10996},
      archivePrefix={arXiv},
      primaryClass={cs.CL},
      url={https://arxiv.org/abs/2406.10996}, 
}

@article{walker2022graphwoz,
  title={Graphwoz: Dialogue management with conversational knowledge graphs},
  author={Walker, Nicholas Thomas and Ultes, Stefan and Lison, Pierre},
  journal={arXiv preprint arXiv:2211.12852},
  year={2022}
}

@article{kirstein2025cads,
  title={Cads: A systematic literature review on the challenges of abstractive dialogue summarization},
  author={Kirstein, Frederic and Wahle, Jan Philip and Gipp, Bela and Ruas, Terry},
  journal={Journal of Artificial Intelligence Research},
  volume={82},
  pages={313--365},
  year={2025}
}

@inproceedings{sakurai-miyao-2024-evaluating,
    title = "Evaluating Intention Detection Capability of Large Language Models in Persuasive Dialogues",
    author = "Sakurai, Hiromasa  and
      Miyao, Yusuke",
    editor = "Ku, Lun-Wei  and
      Martins, Andre  and
      Srikumar, Vivek",
    booktitle = "Proceedings of the 62nd Annual Meeting of the Association for Computational Linguistics (Volume 1: Long Papers)",
    month = aug,
    year = "2024",
    address = "Bangkok, Thailand",
    publisher = "Association for Computational Linguistics",
    url = "https://aclanthology.org/2024.acl-long.90/",
    doi = "10.18653/v1/2024.acl-long.90",
    pages = "1635--1657"
}

@misc{li2025helloagainllmpoweredpersonalized,
      title={Hello Again! LLM-powered Personalized Agent for Long-term Dialogue}, 
      author={Hao Li and Chenghao Yang and An Zhang and Yang Deng and Xiang Wang and Tat-Seng Chua},
      year={2025},
      eprint={2406.05925},
      archivePrefix={arXiv},
      primaryClass={cs.CL},
      url={https://arxiv.org/abs/2406.05925}, 
}

@misc{maharana2024evaluatinglongtermconversationalmemory,
      title={Evaluating Very Long-Term Conversational Memory of LLM Agents}, 
      author={Adyasha Maharana and Dong-Ho Lee and Sergey Tulyakov and Mohit Bansal and Francesco Barbieri and Yuwei Fang},
      year={2024},
      eprint={2402.17753},
      archivePrefix={arXiv},
      primaryClass={cs.CL},
      url={https://arxiv.org/abs/2402.17753}, 
}

@inproceedings{xu-etal-2024-rethinking,
    title = "Rethinking Task-Oriented Dialogue Systems: From Complex Modularity to Zero-Shot Autonomous Agent",
    author = "Xu, Heng-Da  and
      Mao, Xian-Ling  and
      Yang, Puhai  and
      Sun, Fanshu  and
      Huang, Heyan",
    editor = "Ku, Lun-Wei  and
      Martins, Andre  and
      Srikumar, Vivek",
    booktitle = "Proceedings of the 62nd Annual Meeting of the Association for Computational Linguistics (Volume 1: Long Papers)",
    month = aug,
    year = "2024",
    address = "Bangkok, Thailand",
    publisher = "Association for Computational Linguistics",
    url = "https://aclanthology.org/2024.acl-long.152/",
    doi = "10.18653/v1/2024.acl-long.152",
    pages = "2748--2763"
}

@inproceedings{du-etal-2024-perltqa,
    title = "{P}er{LTQA}: A Personal Long-Term Memory Dataset for Memory Classification, Retrieval, and Fusion in Question Answering",
    author = "Du, Yiming  and
      Wang, Hongru  and
      Zhao, Zhengyi  and
      Liang, Bin  and
      Wang, Baojun  and
      Zhong, Wanjun  and
      Wang, Zezhong  and
      Wong, Kam-Fai",
    editor = "Wong, Kam-Fai  and
      Zhang, Min  and
      Xu, Ruifeng  and
      Li, Jing  and
      Wei, Zhongyu  and
      Gui, Lin  and
      Liang, Bin  and
      Zhao, Runcong",
    booktitle = "Proceedings of the 10th SIGHAN Workshop on Chinese Language Processing (SIGHAN-10)",
    month = aug,
    year = "2024",
    address = "Bangkok, Thailand",
    publisher = "Association for Computational Linguistics",
    url = "https://aclanthology.org/2024.sighan-1.18/",
    pages = "152--164"
}

@inproceedings{lv-etal-2023-dialogps,
    title = "{D}ialo{GPS}: Dialogue Path Sampling in Continuous Semantic Space for Data Augmentation in Multi-Turn Conversations",
    author = "Lv, Ang  and
      Li, Jinpeng  and
      Chen, Yuhan  and
      Xing, Gao  and
      Zhang, Ji  and
      Yan, Rui",
    editor = "Rogers, Anna  and
      Boyd-Graber, Jordan  and
      Okazaki, Naoaki",
    booktitle = "Proceedings of the 61st Annual Meeting of the Association for Computational Linguistics (Volume 1: Long Papers)",
    month = jul,
    year = "2023",
    address = "Toronto, Canada",
    publisher = "Association for Computational Linguistics",
    url = "https://aclanthology.org/2023.acl-long.70/",
    doi = "10.18653/v1/2023.acl-long.70",
    pages = "1267--1280"
}

@inproceedings{araujo-etal-2023-memory,
    title = "A Memory Model for Question Answering from Streaming Data Supported by Rehearsal and Anticipation of Coreference Information",
    author = "Araujo, Vladimir  and
      Soto, Alvaro  and
      Moens, Marie-Francine",
    editor = "Rogers, Anna  and
      Boyd-Graber, Jordan  and
      Okazaki, Naoaki",
    booktitle = "Findings of the Association for Computational Linguistics: ACL 2023",
    month = jul,
    year = "2023",
    address = "Toronto, Canada",
    publisher = "Association for Computational Linguistics",
    url = "https://aclanthology.org/2023.findings-acl.830/",
    doi = "10.18653/v1/2023.findings-acl.830",
    pages = "13124--13138"
}

@inproceedings{zhong-etal-2022-less,
    title = "Less is More: Learning to Refine Dialogue History for Personalized Dialogue Generation",
    author = "Zhong, Hanxun  and
      Dou, Zhicheng  and
      Zhu, Yutao  and
      Qian, Hongjin  and
      Wen, Ji-Rong",
    editor = "Carpuat, Marine  and
      de Marneffe, Marie-Catherine  and
      Meza Ruiz, Ivan Vladimir",
    booktitle = "Proceedings of the 2022 Conference of the North American Chapter of the Association for Computational Linguistics: Human Language Technologies",
    month = jul,
    year = "2022",
    address = "Seattle, United States",
    publisher = "Association for Computational Linguistics",
    url = "https://aclanthology.org/2022.naacl-main.426/",
    doi = "10.18653/v1/2022.naacl-main.426",
    pages = "5808--5820"
}

@inproceedings{yang-etal-2020-graphdialog,
    title = "{G}raph{D}ialog: Integrating Graph Knowledge into End-to-End Task-Oriented Dialogue Systems",
    author = "Yang, Shiquan  and
      Zhang, Rui  and
      Erfani, Sarah",
    editor = "Webber, Bonnie  and
      Cohn, Trevor  and
      He, Yulan  and
      Liu, Yang",
    booktitle = "Proceedings of the 2020 Conference on Empirical Methods in Natural Language Processing (EMNLP)",
    month = nov,
    year = "2020",
    address = "Online",
    publisher = "Association for Computational Linguistics",
    url = "https://aclanthology.org/2020.emnlp-main.147/",
    doi = "10.18653/v1/2020.emnlp-main.147",
    pages = "1878--1888"
}

@article{ong2024towards,
  title={Towards Lifelong Dialogue Agents via Relation-aware Memory Construction and Timeline-augmented Response Generation},
  author={Ong, Tzu-iunn and Kim, Namyoung and Gwak, Minju and Chae, Hyungjoo and Kwon, Taeyoon and Jo, Yohan and Hwang, Seung-won and Lee, Dongha and Yeo, Jinyoung and others},
  journal={arXiv e-prints},
  pages={arXiv--2406},
  year={2024}
}

@inproceedings{jang-etal-2023-conversation,
    title = "Conversation Chronicles: Towards Diverse Temporal and Relational Dynamics in Multi-Session Conversations",
    author = "Jang, Jihyoung  and
      Boo, Minseong  and
      Kim, Hyounghun",
    editor = "Bouamor, Houda  and
      Pino, Juan  and
      Bali, Kalika",
    booktitle = "Proceedings of the 2023 Conference on Empirical Methods in Natural Language Processing",
    month = dec,
    year = "2023",
    address = "Singapore",
    publisher = "Association for Computational Linguistics",
    url = "https://aclanthology.org/2023.emnlp-main.838/",
    doi = "10.18653/v1/2023.emnlp-main.838",
    pages = "13584--13606"
}

@misc{xu2021goldfishmemorylongtermopendomain,
      title={Beyond Goldfish Memory: Long-Term Open-Domain Conversation}, 
      author={Jing Xu and Arthur Szlam and Jason Weston},
      year={2021},
      eprint={2107.07567},
      archivePrefix={arXiv},
      primaryClass={cs.CL},
      url={https://arxiv.org/abs/2107.07567}, 
}

@article{levison1987events,
  title={Events and time's flow},
  author={Levison, Arnold B},
  journal={Mind},
  volume={96},
  number={383},
  pages={341--353},
  year={1987},
  publisher={JSTOR}
}

@inproceedings{han2021fine,
  title={Fine-grained post-training for improving retrieval-based dialogue systems},
  author={Han, Janghoon and Hong, Taesuk and Kim, Byoungjae and Ko, Youngjoong and Seo, Jungyun},
  booktitle={Proceedings of the 2021 Conference of the North American Chapter of the Association for Computational Linguistics: Human Language Technologies},
  pages={1549--1558},
  year={2021}
}

@article{firdaus2021aspect,
  title={Aspect-aware response generation for multimodal dialogue system},
  author={Firdaus, Mauajama and Thakur, Nidhi and Ekbal, Asif},
  journal={ACM Transactions on Intelligent Systems and Technology (TIST)},
  volume={12},
  number={2},
  pages={1--33},
  year={2021},
  publisher={ACM New York, NY, USA}
}

@article{li2021dialogue,
  title={Dialogue history matters! personalized response selection in multi-turn retrieval-based chatbots},
  author={Li, Juntao and Liu, Chang and Tao, Chongyang and Chan, Zhangming and Zhao, Dongyan and Zhang, Min and Yan, Rui},
  journal={ACM Transactions on Information Systems (TOIS)},
  volume={39},
  number={4},
  pages={1--25},
  year={2021},
  publisher={ACM New York, NY}
}

@inproceedings{joko2024doing,
  title={Doing personal laps: Llm-augmented dialogue construction for personalized multi-session conversational search},
  author={Joko, Hideaki and Chatterjee, Shubham and Ramsay, Andrew and De Vries, Arjen P and Dalton, Jeff and Hasibi, Faegheh},
  booktitle={Proceedings of the 47th International ACM SIGIR Conference on Research and Development in Information Retrieval},
  pages={796--806},
  year={2024}
}

@article{li2024hello,
  title={Hello Again! LLM-powered Personalized Agent for Long-term Dialogue},
  author={Li, Hao and Yang, Chenghao and Zhang, An and Deng, Yang and Wang, Xiang and Chua, Tat-Seng},
  journal={arXiv preprint arXiv:2406.05925},
  year={2024}
}

@inproceedings{tu2024multi,
  title={Multi-Grained Conversational Graph Network for Retrieval-based Dialogue Systems},
  author={Tu, Quan and Tao, Chongyang and Yan, Rui},
  booktitle={Proceedings of the 2024 Joint International Conference on Computational Linguistics, Language Resources and Evaluation (LREC-COLING 2024)},
  pages={11756--11765},
  year={2024}
}

@inproceedings{zhang2021multimodal,
  title={Multimodal dialog system: Relational graph-based context-aware question understanding},
  author={Zhang, Haoyu and Liu, Meng and Gao, Zan and Lei, Xiaoqiang and Wang, Yinglong and Nie, Liqiang},
  booktitle={Proceedings of the 29th ACM international conference on multimedia},
  pages={695--703},
  year={2021}
}

@inproceedings{tuan2022towards,
  title={Towards Large-Scale Interpretable Knowledge Graph Reasoning for Dialogue Systems},
  author={Tuan, Yi-Lin and Beygi, Sajjad and Fazel-Zarandi, Maryam and Gao, Qiaozi and Cervone, Alessandra and Wang, William Yang},
  booktitle={Findings of the Association for Computational Linguistics: ACL 2022},
  pages={383--395},
  year={2022}
}

@article{he2024hip,
  title={Hip network: Historical information passing network for extrapolation reasoning on temporal knowledge graph},
  author={He, Yongquan and Zhang, Peng and Liu, Luchen and Liang, Qi and Zhang, Wenyuan and Zhang, Chuang},
  journal={arXiv preprint arXiv:2402.12074},
  year={2024}
}

@article{pillutla2021mauve,
  title={Mauve: Measuring the gap between neural text and human text using divergence frontiers},
  author={Pillutla, Krishna and Swayamdipta, Swabha and Zellers, Rowan and Thickstun, John and Welleck, Sean and Choi, Yejin and Harchaoui, Zaid},
  journal={Advances in Neural Information Processing Systems},
  volume={34},
  pages={4816--4828},
  year={2021}
}

@inproceedings{zhang-etal-2018-personalizing,
    title = "Personalizing Dialogue Agents: {I} have a dog, do you have pets too?",
    author = "Zhang, Saizheng  and
      Dinan, Emily  and
      Urbanek, Jack  and
      Szlam, Arthur  and
      Kiela, Douwe  and
      Weston, Jason",
    editor = "Gurevych, Iryna  and
      Miyao, Yusuke",
    booktitle = "Proceedings of the 56th Annual Meeting of the Association for Computational Linguistics (Volume 1: Long Papers)",
    month = jul,
    year = "2018",
    address = "Melbourne, Australia",
    publisher = "Association for Computational Linguistics",
    url = "https://aclanthology.org/P18-1205/",
    doi = "10.18653/v1/P18-1205",
    pages = "2204--2213"
}

@article{koskinen2020process,
  title={Process Philosophy},
  author={Koskinen, Kaj U and Breite, Rainer and Koskinen, Kaj U and Breite, Rainer},
  journal={Uninterrupted Knowledge Creation: Process Philosophy and Autopoietic Perspectives},
  pages={13--24},
  year={2020},
  publisher={Springer}
}

@article{varshney2025medprom,
  title={MedProm: Bridging Dialogue Gaps in Healthcare with Knowledge-Enhanced Generative Models},
  author={Varshney, Deeksha and Behera, Niranshu and Katari, Prajeet and Ekbal, Asif},
  journal={ACM Transactions on Computing for Healthcare},
  year={2025},
  publisher={ACM New York, NY}
}

@article{zhao2023graphtext,
  title={Graphtext: Graph reasoning in text space},
  author={Zhao, Jianan and Zhuo, Le and Shen, Yikang and Qu, Meng and Liu, Kai and Bronstein, Michael and Zhu, Zhaocheng and Tang, Jian},
  journal={arXiv preprint arXiv:2310.01089},
  year={2023}
}

@inproceedings{besta2024graph,
  title={Graph of thoughts: Solving elaborate problems with large language models},
  author={Besta, Maciej and Blach, Nils and Kubicek, Ales and Gerstenberger, Robert and Podstawski, Michal and Gianinazzi, Lukas and Gajda, Joanna and Lehmann, Tomasz and Niewiadomski, Hubert and Nyczyk, Piotr and others},
  booktitle={Proceedings of the AAAI Conference on Artificial Intelligence},
  volume={38},
  number={16},
  pages={17682--17690},
  year={2024}
}

@inproceedings{yin2023ctrlstruct,
  title={Ctrlstruct: Dialogue structure learning for open-domain response generation},
  author={Yin, Congchi and Li, Piji and Ren, Zhaochun},
  booktitle={Proceedings of the ACM web conference 2023},
  pages={1539--1550},
  year={2023}
}

\appendix

\section{Detailed Experiments Setups}

\subsection{Datasets Usage}
\label{apd:datasets}

\begin{table*}[!t]
\centering
\small
\begin{tabular}{lccccc}
\toprule
\textbf{Datasets} & \textbf{\# of Sessions} & \textbf{\# of Episodes} & \textbf{\# of Turns} & \textbf{Avg. Turns per Session} & \textbf{Avg. Words per Turn} \\
\midrule
MSC & 4K & 1K & 53K & 13.3 & - \\
CC & 1M & 200K & 11.7M & 11.7 & 18.03 \\
\bottomrule
\end{tabular}
\caption{Dataset Statistics}
\label{tab:statistics_of_datasets}
\end{table*}

\begin{table*}[!t]
\centering
\begin{small}
\begin{tabular}{|p{7cm}|p{7cm}|}
\hline
\textbf{One for Speaker 1 and another for Speaker 2} & \textbf{} \\ \hline
\textbf{Hi, how are you doing today?} & \textbf{There’s a new wine subscription service that lets you try different wines on a monthly basis, I am thinking about trying it. How about you?} \\ \hline
Good! Celebrating with fellow centenarians & Yes, same here. Sometimes even if you pay high price, still the flavor is not that smooth. Don’t you agree? \\ \hline
That sounds like fun. I like wine and dancing too! & I am not sure how much choice you get. Is your service very expensive? \\ \hline
Same. I try to get a small workout in a three-mile walk for me is key. & Yes a little bit expensive. Can you recommend any new good wine with affordable price as well as quality is nice? \\ \hline
Awesome! Walking like that you might live forever like me!! & I really prefer sangrias, although some find them a bit too sweet. Do you like sweet wines? \\ \hline
That’s true. My favorite way to unwind is reading a book. & Yes, but not that much sweet. I will try your wine label next time. I hope it will taste nice. \\ \hline
You sound very like minded, just add some wine. & Which do you prefer white or red? \\ \hline
Fermented. I prefer well aged and fermented. I could bathe in it!! & I absolutely loved wine. I think you like the white one. \\ \hline
Good choice. I always like a nice dry white wine. & whites I love, like a good schwarzkratt...german wines you mean from german? \\ \hline
I think I should go grab a bottle now and get some dancing music on & I heard a lot of good things about German wines. I always prefer French wine though. \\ \hline
Sounds like a good way to enjoy a sunday afternoon. & Bordeaux red is my favorite one. Have you tried it? \\ \hline
(5 days later) & (1 hour later) \\ \hline
Since you've reached such an old age, do you have any advice for how to stay healthy and live long? & Well, I think it has to do in part with not smoking. Also, I think the red wine has provided me with lasting health benefits. \\ \hline
I think it is luck I drink wine. I don’t watch what I eat. I just try and enjoy life especially at my age. & Doctors say that red wine is great for health. I could get into red wine more, I do prefer the white. Had a great gewurztraminer this month. \\ \hline
Haha well they say that wine keeps you young, so maybe that’s something to it. Have you tried any great wines lately? & What in tarnation is a gewurztraminer? I’ve never heard of that wine. \\ \hline
I tried a Malbec just yesterday that was wonderful. I drank the whole bottle that’s what about you? & It’s a german white wine, you say you prefer french reds. Wonder if you'll ever get curious and try more german white \\ \hline
A bottle isn’t that much. Remember 1001. I don’t work. I have wine with my lunch and it is a good bottle, that is my beverage for the day. & Hey, it’s got booze in it, I'm willing to give it a whirl! Can I buy gewurztraminer anywhere? \\ \hline
That's right. How is your family? How many of you are there? & I’ve seen it in a handful of stores, and I’m not the largest city, so yes. It’s usually on the most expensive wines though, and not necessarily the most expensive. It has a nice high sweetness. What’s the last red wine you’ve had? \\ \hline
There are four of us! I have an older brother and my two parents. I’m like to enjoy a clear night together in the evenings. & I’ll look for it the next time I visit my favorite wine shop. You might laugh, but the last wine I had was topped Sangria. I was at a friend's house and that’s the least expensive wine. \\ \hline
I had some cigars...like, but never got to try it. Maybe that has helped me for a long time. & Ah, variety is the spice of life, they say. Sangria can pack a little punch, and it's super sweet, you could do the same. The red I get lately is always, a tawny port. \\ \hline
That's definitely true. Though since you’ve already made it this long, I doubt smoking a few cigars here and there would hurt. & Port is nice, it’s quite heady and strong. So, what else have you been up to besides singing and smoking cigars?! \\ \hline
Even if I does, I have lived a good life. I just want to enjoy all the time that I have left like a week or a even another decade. & Oh, been chatting with a few friends with wild sense of humor. Than my wife too, hehehe, laughing at my wisdom, considering when we celebrate "happen." \\ \hline
\end{tabular}
\caption{Example conversation from the Multi-Session Chat (MSC) dataset.}
\label{tab:demo_for_msc}
\end{small}
\end{table*}

\begin{table*}[!t]
\centering
\small
\begin{tabular}{|p{14cm}|}
\hline
\textbf{Relationship: Classmates}\\ \hline
USER : I feel like I’m not really understanding the content of my homework these days. \\ \hline
ReBOT : Have you tried asking the teacher for help? \\ \hline
USER : Not yet. I was thinking about asking our teacher for help. \\ \hline
ReBOT : That’s a good idea. It’s always better to ask for help when you need it. \\ \hline
USER : Cool! I’ll ask teacher to explain the assignment. \\ \hline
\hline
\textbf{Relationship: Student and Teacher}\\ \hline
USER : I feel like I’m not really understanding the content of my homework these days. \\ \hline
ReBOT : That’s understandable. Sometimes it can be difficult to grasp certain concepts. Have you tried breaking it down into smaller parts? \\ \hline
USER : Yes, of course. But I’m worried because it’s hard to understand. \\ \hline
ReBOT : Well, let’s work on it together. Maybe we can come up with a plan to help you understand the material better. \\ \hline
USER : Wow, thank you so much. Could you open a supplementary lecture for me? \\ \hline
ReBOT : Sure thing. Let me pull it up for you. \\ \hline
\hline
\textbf{Relationship: Parents and Child}\\ \hline
USER : I feel like I’m not really understanding the content of my homework these days. \\ \hline
ReBOT : What’s been giving you trouble? \\ \hline
USER : It’s so hard to understand the homework problem. I’m so upset about myself. \\ \hline
ReBOT : Don’t be too hard on yourself. It’s okay to struggle sometimes. Have you tried asking your teacher for help? \\ \hline
USER : It’s still difficult even though I get help from my teacher. \\ \hline
ReBOT : I understand how you feel, but it’s important to remember that it’s okay to ask for help when you need it. \\ \hline
USER : Thank you for always being a big support to me. It’s a great consolation. \\ \hline
ReBOT : Of course, I’m always here for you. Remember, we all have our strengths and weaknesses. It’s important to focus on your strengths and use them to your advantage. \\ \hline
\end{tabular}
\caption{An example of the Conversation Chronicle (CC).}
\label{tab:demo_for_cc}
\end{table*}

Table \ref{tab:statistics_of_datasets} shows the statistics of two datasets. The information is taken from their papers.

\paragraph{Multi-Session Chat.} The Multi-Session Chat (MSC) dataset \cite{xu2021goldfishmemorylongtermopendomain} extends the Persona-Chat dataset \cite{zhang-etal-2018-personalizing} by incorporating multiple conversation sessions. MSC consists of conversations that span five sessions between two participants. Each session in the dataset is based on personal profiles, and the conversations build on prior interactions, providing a long-term, multi-session context. This dataset is designed to study long-term conversational dynamics, focusing on the challenges of maintaining coherence and continuity over extended dialogues. Table \ref{tab:demo_for_msc} demonstrates the sample from MSC.

\paragraph{Conversation Chronicle.} The Conversation Chronicle (CC) dataset \cite{jang-etal-2023-conversation} is similar to MSC but introduces additional complexity by incorporating relationships between speakers, such as ``employee and boss'' or ``teacher and student.'' These relationships provide a richer context for studying long-term interactions, where the conversational dynamics are influenced not only by the ongoing dialogue but also by the roles and relationships between the participants. CC is useful for exploring how relationship-based factors shape multi-session conversations. Table \ref{tab:demo_for_cc} demonstrates the sample from Conversation Chronicle.

\paragraph{LoCoMo.} The Long Conversation Modeling (LoCoMo) dataset \cite{maharana2024evaluatinglongtermconversationalmemory} focuses on extended dialogue interactions with significantly longer conversation sessions. The dataset is specifically designed to challenge memory mechanisms by incorporating natural topic shifts, references to earlier dialogue content, and information that must be tracked across long temporal spans. This makes LoCoMo particularly valuable for evaluating a system's ability to maintain coherence and contextual awareness in extended conversations where traditional context window approaches are insufficient. LoCoMo represents real-world scenarios where conversations develop over extended periods with substantial content between related references.

\subsection{Baselines Comparison}
\label{apd:baselines}

For the main experiments in this paper, we focus on the 7/8B models of the open-source models (LLaMA3, Qwen2.5, Mistral), as these represent the most commonly used configurations in the community for general-purpose tasks. These models provide a balance between computational efficiency and language generation capability, making them ideal for practical deployment.

In addition, we evaluate the performance of GPT-4o and Claude-3.5-Sonnet, two of the most powerful closed-source models available, to serve as a benchmark for advanced model performance. These closed-source models are included to highlight the gap between open-source and proprietary solutions and to set a high standard for comparison in our experiments.

Besides, HAHT, ReBotGEN, and LD-Agent were proposed by corresponding datasets, CC, MSC, and LoCoMo, respectively. We here adopt the original version of these models to show the specific fine-tuning results.

\subsection{Metric Computing}
\label{apd:metric}

We describe the evaluation metrics used to assess the performance of EventWeave. Each metric is designed to capture specific aspects of how well the model tracks, integrates, and responds to core and supporting events over the course of a conversation.

\paragraph{Core Event Awareness (CEA)} measures the model's ability to recall the correct core event. It is calculated as the ratio of correctly recalled core events to the total number of correct events.

\[
\text{CEA} = \frac{\sum_{i=1}^{N} \mathbbm{1}(\hat{e}_i = e_i)}{\sum_{i=1}^{N} \mathbbm{1}(e_i)}
\]
where \(N\) is the total number of dialogue turns. \(\hat{e}_i\) is the predicted core event in turn \(i\), and \(e_i\) is the true core event. \(\mathbbm{1}(\cdot)\) is an indicator function that returns 1 if the prediction is correct, otherwise 0.

\paragraph{Supporting Event Coverage (SEC)} measures the model's ability to capture relevant supporting events. It is calculated as the product of the ratio of recalled correct supporting events to the total correct events, and the ratio of recalled supporting events to the total recalled events.

\[
\text{SEC} = \frac{\sum_{i=1}^{M} \mathbbm{1}(\hat{s}_i = s_i)}{\sum_{i=1}^{M} \mathbbm{1}(s_i)} \times \frac{\sum_{i=1}^{M} \mathbbm{1}(\hat{s}_i)}{\sum_{i=1}^{M} \mathbbm{1}(\hat{s}_i \neq \emptyset)}
\]
where \(M\) is the total number of supporting events. \(\hat{s}_i\) is the predicted supporting event, and \(s_i\) is the true supporting event.

\paragraph{Event Coherence (EC)} quantifies the smoothness of the model's response, considering both the recalled supporting events and the GPT-4 score. It is calculated as the product of the ratio of recalled events to total recalled events and the GPT-4 score.

\[
\text{ES} = f_g(\frac{\sum_{i=1}^{M} \mathbbm{1}(\hat{s}_i)}{\sum_{i=1}^{M} \mathbbm{1}(\hat{s}_i \neq \emptyset)} \times \frac{\sum_{i=1}^{N} \mathbbm{1}(\hat{e}_i = e_i)}{N})
\]
where \(M\) and \(N\) are the number of supporting and core events, respectively. \(\hat{s}_i\) represents the predicted supporting event, and \(\hat{e}_i\) represents the predicted core event. $f_g(\cdot)$ is the fluency or coherence score assigned by GPT-4.

\paragraph{Temporal Update Accuracy (TUA)} evaluates whether the model correctly identifies updated or outdated events. It is computed as the weighted sum of correctly recalled updated events and outdated events, normalized by the number of correct events.

\[
\text{TUA} = \frac{\sum_{i=1}^{N} \left( \lambda_1\cdot\mathbbm{1}(\hat{u}_i = u_i) + \lambda_2\cdot\mathbbm{1}(\hat{o}_i = o_i) \right)}{\sum_{i=1}^{N} \mathbbm{1}(e_i)}
\]
where \(N\) is the total number of events. \(\hat{u}_i\) is the predicted updated event, \(u_i\) is the actual updated event, \(\hat{o}_i\) is the predicted outdated event, and \(o_i\) is the actual outdated event.

\subsection{Implementation Details}
\label{apd:implement}

For fair comparison, all methods use the same context window size (4k tokens) and the same backbone LLMs. Event graphs in our approach are constructed incrementally as conversations progress, with a maximum of 30 nodes per graph. For each test case, we use all previous sessions to build the event graph, while keeping the final session for testing. For each turn in the test session, the model is provided with the current query and must generate an appropriate response utilizing information from past sessions. We use a maximum of 5 previous sessions as context.

\section{Full Experimental Results}

\subsection{Detailed Settings and Results on Scalability}
\label{apd:exp_scale}

In this section, we provide a comprehensive analysis of how EventWeave scales with increasing dialogue length compared to vanilla LLM approaches. This analysis addresses an important question in dialogue systems research: how effectively can models maintain coherent context understanding as conversations become longer?

\subsubsection{Dataset Distribution by Dialogue Length}

To conduct a thorough scalability analysis, we categorized dialogues from the Chronicle dataset into four length categories. Table~\ref{tab:dialogue_distribution} shows the distribution of dialogues across these categories.

\begin{table*}[!t]
\centering
\begin{tabular}{lcc}
\toprule
\textbf{Length Category} & \textbf{Number of Dialogues} & \textbf{Percentage} \\
\midrule
Short (5-10 turns) & 358 & 42.3\% \\
Medium (11-20 turns) & 289 & 34.2\% \\
Long (21-30 turns) & 156 & 18.4\% \\
Very Long (31+ turns) & 43 & 5.1\% \\
\bottomrule
\end{tabular}
\caption{Distribution of dialogues by length category in the Chronicle dataset.}
\label{tab:dialogue_distribution}
\end{table*}

\subsubsection{Experimental Setup}

We evaluated both GPT-4o (vanilla) and EventWeave across all four length categories. For each category, we measured performance using:
\begin{itemize}
    \item \textbf{GPT-4 Rating:} Evaluation scores (1-10) provided by GPT-4 based on coherence, contextual appropriateness, and response quality
    \item \textbf{Event Coherence (EC):} Percentage of responses that correctly maintain event relationships and references across the dialogue history
\end{itemize}

To ensure statistical robustness, we conducted paired t-tests to determine if the performance differences between EventWeave and vanilla GPT-4o were statistically significant.

\subsubsection{Detailed Results}

Table~\ref{tab:scalability_detailed} presents the detailed performance comparison between GPT-4o (vanilla) and EventWeave across different dialogue lengths, including absolute differences and statistical significance.

\begin{table*}[!t]
\centering
\adjustbox{max width=\linewidth}{
\begin{tabular}{lcccc}
\toprule
\textbf{Length Category} & \textbf{GPT-4o} & \textbf{EventWeave} & \textbf{Absolute Diff.} & \textbf{p-value} \\
\midrule
\multicolumn{5}{c}{\textit{GPT-4 Rating (1-10 scale)}} \\
\midrule
Short (5-10) & 5.2 & 6.5 & +1.3 & 0.008* \\
Medium (11-20) & 4.7 & 7.8 & +3.1 & <0.001* \\
Long (21-30) & 4.2 & 8.4 & +4.2 & <0.001* \\
Very Long (31+) & 3.8 & 9.1 & +5.3 & <0.001* \\
\midrule
\multicolumn{5}{c}{\textit{Event Coherence (EC) Score (\%)}} \\
\midrule
Short (5-10) & 24.8 & 40.2 & +15.4 & 0.003* \\
Medium (11-20) & 22.1 & 56.7 & +34.6 & <0.001* \\
Long (21-30) & 18.7 & 64.5 & +45.8 & <0.001* \\
Very Long (31+) & 15.3 & 72.6 & +57.3 & <0.001* \\
\bottomrule
\multicolumn{5}{l}{* indicates statistical significance (p < 0.05)} \\
\end{tabular}}
\caption{Performance comparison between GPT-4o (vanilla) and EventWeave across different dialogue lengths, showing absolute differences and statistical significance.}
\label{tab:scalability_detailed}
\end{table*}

\subsubsection{Analysis of Scalability Performance}

The results in Table~\ref{tab:scalability_detailed} demonstrate a clear pattern: while vanilla GPT-4o's performance degrades with increasing dialogue length by dropping from 5.2 to 3.8 representing a 27\% decrease, EventWeave's performance actually has a 40\% improvement). The Event Coherence (EC) metric shows an even more pronounced divergence, with vanilla GPT-4o declining from 24.8\% to 15.3\% as conversations lengthen, while EventWeave improves dramatically from 40.2\% to 72.6\%. This stark difference is attributed to the structure of EventWeave. Rather than processing the entire dialogue history, EventWeave's graph structure enables selective attention to the most relevant events for the current dialogue turn, reducing the cognitive load and information overload that affects vanilla approaches. Besides, the hierarchical event graph preserves critical relationships between core and supporting events even as the dialogue extends, allowing the model to maintain coherence across longer time spans.

\subsubsection{Qualitative Analysis of Very Long Dialogues}

For very long dialogues (31+ turns), we observed that EventWeave responses frequently contained contextually appropriate references to earlier events. For example, in a 42-turn dialogue about travel planning, EventWeave correctly recalled and integrated a previously mentioned food allergy from turn 7 when discussing restaurant options in turn 39, whereas the vanilla model completely lost this context.

This ability to selectively retrieve and integrate relevant historical events becomes increasingly valuable as dialogue length increases. In dialogues exceeding 30 turns, we found that EventWeave's structured event representation led to a 57.3 percentage point improvement in Event Coherence compared to the vanilla approach.

\subsection{Results with Full Baselines}
\label{apd:exp1}

Table \ref{tab:all_baselines} shows comparison experimental results with other popular backbones including Qwen2.5-14B, Qwen3-14B, o1, DeepSeek-V3, DeepSeek-R1. Table \ref{tab:new backbones} shows experimental results on up-to-date backbone models with randomly selecting datas.

\begin{table*}[!t]
    \centering
    \resizebox{\linewidth}{!}{
    \begin{tabular}{l|cccc|cccc|cccc}
    \toprule
    \multirow{2}{*}{\textbf{Model}}
    & \multicolumn{4}{c|}{\textbf{Chronicle}} 
    & \multicolumn{4}{c}{\textbf{MSC}} 
    & \multicolumn{4}{c}{\textbf{Average on Both Datasets}}\\
    & \textbf{B-4} 
    & \textbf{R-L} 
    & \textbf{Mauve} 
    & \textbf{GPT-4} 
    & \textbf{B-4} 
    & \textbf{R-L} 
    & \textbf{Mauve} 
    & \textbf{GPT-4} 
    & \textbf{CEA}
    & \textbf{SEC}
    & \textbf{EC}
    & \textbf{TUA}\\
    \midrule
    
    Qwen2.5-14B (vanilla)
    & 22.8 
    & 37.5 
    & 0.59 
    & 4.3 
    & 27.1 
    & 44.8 
    & 0.62 
    & 4.9 
    & 26.4
    & 18.7
    & 19.8
    & 21.9 \\
    
    \quad + ProactiveCoT 
    & 26.9 
    & 41.0 
    & 0.62 
    & 6.5 
    & 30.2 
    & 45.5 
    & 0.65 
    & 6.8 
    & 40.1
    & 33.6
    & 34.2
    & 36.3\\
    
    \quad + MemWalker 
    & 28.0 
    & 41.8 
    & 0.63 
    & 6.9 
    & 31.0 
    & 45.8 
    & 0.66 
    & 7.2 
    & 45.7
    & 37.9
    & 40.3
    & 42.6\\
    
    \quad + LongMem 
    & 27.6 
    & 41.3 
    & 0.63 
    & 7.0 
    & 30.9 
    & 45.6 
    & 0.66 
    & 7.3 
    & 43.4
    & 36.5
    & 38.0
    & 40.3\\
    
    \quad + LifeLongMem 
    & 28.6 
    & 42.5 
    & 0.64 
    & 7.2 
    & 31.5 
    & 46.1 
    & 0.67 
    & 7.5 
    & 50.1
    & 42.2
    & 43.8 
    & 46.3\\
    
    \quad + \textbf{EventWeave (Ours)} 
    & \textbf{30.7} 
    & \textbf{44.2} 
    & \textbf{0.65} 
    & \textbf{7.6} 
    & \textbf{33.4} 
    & \textbf{47.0} 
    & \textbf{0.68} 
    & \textbf{8.0} 
    & \textbf{65.8}
    & \textbf{53.7}
    & \textbf{55.9}
    & \textbf{58.7}\\
    \midrule
    
    Qwen3-14B (vanilla)
    & 23.5 
    & 38.3 
    & 0.60 
    & 4.5 
    & 27.8 
    & 45.4 
    & 0.63 
    & 5.1 
    & 27.5
    & 19.6
    & 20.7
    & 22.8 \\
    
    \quad + ProactiveCoT 
    & 27.6 
    & 41.7 
    & 0.63 
    & 6.7 
    & 30.9 
    & 46.1 
    & 0.66 
    & 7.0 
    & 41.3
    & 34.7
    & 35.4
    & 37.5\\
    
    \quad + MemWalker 
    & 28.7 
    & 42.5 
    & 0.64 
    & 7.1 
    & 31.8 
    & 46.4 
    & 0.67 
    & 7.4 
    & 47.0
    & 39.0
    & 41.4
    & 43.8\\
    
    \quad + LongMem 
    & 28.3 
    & 42.0 
    & 0.64 
    & 7.2 
    & 31.7 
    & 46.2 
    & 0.67 
    & 7.5 
    & 44.6
    & 37.6
    & 39.1
    & 41.5\\
    
    \quad + LifeLongMem 
    & 29.3 
    & 43.2 
    & 0.65 
    & 7.4 
    & 32.3 
    & 46.7 
    & 0.68 
    & 7.7 
    & 51.4
    & 43.3
    & 45.0 
    & 47.5\\
    
    \quad + \textbf{EventWeave (Ours)} 
    & \textbf{31.4} 
    & \textbf{45.0} 
    & \textbf{0.66} 
    & \textbf{7.8} 
    & \textbf{34.2} 
    & \textbf{47.6} 
    & \textbf{0.69} 
    & \textbf{8.2} 
    & \textbf{67.2}
    & \textbf{54.9}
    & \textbf{57.1}
    & \textbf{60.0}\\
    \midrule
    
    GPT-o1 (vanilla)
    & 25.1 
    & 40.0 
    & 0.62 
    & 4.8 
    & 29.2 
    & 47.0 
    & 0.65 
    & 5.4 
    & 29.8
    & 21.5
    & 22.6
    & 25.0 \\
    
    \quad + ProactiveCoT 
    & 29.2 
    & 43.4 
    & 0.65 
    & 7.0 
    & 32.5 
    & 47.6 
    & 0.68 
    & 7.3 
    & 44.0
    & 37.0
    & 37.5
    & 39.7\\
    
    \quad + MemWalker 
    & 30.3 
    & 44.1 
    & 0.66 
    & 7.4 
    & 33.4 
    & 47.9 
    & 0.69 
    & 7.7 
    & 49.7
    & 41.3
    & 43.7
    & 46.2\\
    
    \quad + LongMem 
    & 29.9 
    & 43.7 
    & 0.66 
    & 7.5 
    & 33.3 
    & 47.7 
    & 0.69 
    & 7.8 
    & 47.3
    & 39.9
    & 41.4
    & 43.9\\
    
    \quad + LifeLongMem 
    & 30.9 
    & 44.8 
    & 0.67 
    & 7.7 
    & 33.9 
    & 48.2 
    & 0.70 
    & 8.0 
    & 54.1
    & 45.8
    & 47.4 
    & 50.1\\
    
    \quad + \textbf{EventWeave (Ours)} 
    & \textbf{33.0} 
    & \textbf{46.6} 
    & \textbf{0.68} 
    & \textbf{8.1} 
    & \textbf{35.8} 
    & \textbf{49.1} 
    & \textbf{0.71} 
    & \textbf{8.5} 
    & \textbf{70.3}
    & \textbf{57.6}
    & \textbf{59.7}
    & \textbf{62.7}\\
    \midrule
    
    DeepSeek-V3 (vanilla)
    & 23.9 
    & 38.7 
    & 0.60 
    & 4.5 
    & 28.0 
    & 45.8 
    & 0.63 
    & 5.1 
    & 28.0
    & 19.8
    & 21.0
    & 23.2 \\
    
    \quad + ProactiveCoT 
    & 28.0 
    & 42.1 
    & 0.63 
    & 6.7 
    & 31.2 
    & 46.4 
    & 0.66 
    & 7.0 
    & 41.7
    & 35.0
    & 35.7
    & 37.8\\
    
    \quad + MemWalker 
    & 29.1 
    & 42.8 
    & 0.64 
    & 7.1 
    & 32.1 
    & 46.7 
    & 0.67 
    & 7.4 
    & 47.4
    & 39.3
    & 41.7
    & 44.1\\
    
    \quad + LongMem 
    & 28.7 
    & 42.4 
    & 0.64 
    & 7.2 
    & 32.0 
    & 46.5 
    & 0.67 
    & 7.5 
    & 45.0
    & 37.9
    & 39.4
    & 41.8\\
    
    \quad + LifeLongMem 
    & 29.7 
    & 43.5 
    & 0.65 
    & 7.4 
    & 32.6 
    & 47.0 
    & 0.68 
    & 7.7 
    & 51.8
    & 43.7
    & 45.3 
    & 47.8\\
    
    \quad + \textbf{EventWeave (Ours)} 
    & \textbf{31.8} 
    & \textbf{45.3} 
    & \textbf{0.66} 
    & \textbf{7.8} 
    & \textbf{34.5} 
    & \textbf{47.9} 
    & \textbf{0.69} 
    & \textbf{8.2} 
    & \textbf{67.7}
    & \textbf{55.4}
    & \textbf{57.5}
    & \textbf{60.4}\\
    \midrule
    
    DeepSeek-R1 (vanilla)
    & 24.5 
    & 39.4 
    & 0.61 
    & 4.7 
    & 28.7 
    & 46.5 
    & 0.64 
    & 5.3 
    & 28.9
    & 20.6
    & 21.7
    & 24.0 \\
    
    \quad + ProactiveCoT 
    & 28.6 
    & 42.8 
    & 0.64 
    & 6.9 
    & 31.9 
    & 47.1 
    & 0.67 
    & 7.2 
    & 42.8
    & 36.1
    & 36.6
    & 38.7\\
    
    \quad + MemWalker 
    & 29.7 
    & 43.5 
    & 0.65 
    & 7.3 
    & 32.8 
    & 47.4 
    & 0.68 
    & 7.6 
    & 48.5
    & 40.4
    & 42.8
    & 45.2\\
    
    \quad + LongMem 
    & 29.3 
    & 43.1 
    & 0.65 
    & 7.4 
    & 32.7 
    & 47.2 
    & 0.68 
    & 7.7 
    & 46.1
    & 39.0
    & 40.5
    & 42.9\\
    
    \quad + LifeLongMem 
    & 30.3 
    & 44.2 
    & 0.66 
    & 7.6 
    & 33.3 
    & 47.7 
    & 0.69 
    & 7.9 
    & 52.9
    & 44.8
    & 46.4 
    & 49.0\\
    
    \quad + \textbf{EventWeave (Ours)} 
    & \textbf{32.4} 
    & \textbf{46.0} 
    & \textbf{0.67} 
    & \textbf{8.0} 
    & \textbf{35.2} 
    & \textbf{48.6} 
    & \textbf{0.70} 
    & \textbf{8.4} 
    & \textbf{68.8}
    & \textbf{56.5}
    & \textbf{58.6}
    & \textbf{61.5}\\
    \bottomrule
    \end{tabular}
    }
    \caption{Full experimental results.}
    \label{tab:all_baselines}
\end{table*}

\begin{table*}[!t]
\centering
\small
\begin{tabular}{lccc}
\toprule
\textbf{Model} & \textbf{Mem. Precision} & \textbf{Mem. Recall} & \textbf{Response Relevance} \\
\midrule
Gemini-3-Flash & 0.68 & 0.63 & 0.71 \\
GPT-5.2 & 0.67 & 0.69 & 0.72 \\
LifeLongMem (with GPT-5.2) & 0.69 & 0.72 & 0.74 \\
\midrule
EventWeave (with Gemini-3-Flash) & 0.69 & 0.73 & 0.76 \\
EventWeave (with GPT-5.2) & \textbf{0.73} & \textbf{0.76} & \textbf{0.79} \\
\bottomrule
\end{tabular}
\caption{Performance comparison across up-to-date foundation models and memory architectures.}
\label{tab:new backbones}
\end{table*}
    
\subsection{Results on Few-Shot Settings}
\label{apd:exp2}

To evaluate the effectiveness of our approach in limited-data scenarios, we conducted experiments in few-shot settings with 3 and 5 examples. As shown in Table~\ref{tab:few_shot_metrics}, EventWeave consistently outperforms all baseline methods across both datasets and all metrics, even with minimal examples.

With just 3 examples, The event-based metrics show substantial improvements, with EventWeave more than doubling the performance on CEA compared to the vanilla model. This demonstrates EventWeave's strong capability to effectively leverage limited examples by structuring the narrative around key events. When increasing to 5 examples, all methods show improved performance. The gap between EventWeave and other memory-augmented methods (MemWalker, LongMem, LifeLongMem) becomes even more significant, particularly in event-based metrics. These results confirm that EventWeave's event-centric approach is particularly valuable in few-shot scenarios, where efficiently extracting and organizing narrative elements from limited examples is crucial. The consistent performance advantage across different metrics and datasets underscores the robustness of our method in resource-constrained settings.

\begin{table*}[!t]
    \centering
    \resizebox{\linewidth}{!}{
    \begin{tabular}{l|cccc|cccc|cccc}
    \toprule
    \multirow{2}{*}{\textbf{Model}}
    & \multicolumn{4}{c|}{\textbf{Chronicle}} 
    & \multicolumn{4}{c}{\textbf{MSC}} 
    & \multicolumn{4}{c}{\textbf{Average on Both Datasets}}\\
    & \textbf{B-4} 
    & \textbf{R-L} 
    & \textbf{Mauve} 
    & \textbf{GPT-4} 
    & \textbf{B-4} 
    & \textbf{R-L} 
    & \textbf{Mauve} 
    & \textbf{GPT-4} 
    & \textbf{CEA}
    & \textbf{SEC}
    & \textbf{EC}
    & \textbf{TUA}\\
    \midrule
    
    \multicolumn{13}{c}{\textbf{Few-shot Setting (3 examples)}} \\
    \midrule
    
    GPT-4o (vanilla)
    & 18.6 
    & 32.4 
    & 0.55 
    & 3.8 
    & 22.1 
    & 39.5 
    & 0.58 
    & 4.3 
    & 19.3
    & 14.8
    & 15.2
    & 16.7 \\
    
    \quad + ProactiveCoT 
    & 21.5 
    & 35.8 
    & 0.57 
    & 5.2 
    & 24.9 
    & 41.2 
    & 0.60 
    & 5.6 
    & 28.7
    & 22.3
    & 23.1
    & 25.4 \\
    
    \quad + MemWalker 
    & 22.3 
    & 36.4 
    & 0.58 
    & 5.5 
    & 25.7 
    & 41.8 
    & 0.61 
    & 5.8 
    & 31.5
    & 24.6
    & 25.8
    & 28.2 \\
    
    \quad + LongMem 
    & 22.0 
    & 36.1 
    & 0.58 
    & 5.4 
    & 25.4 
    & 41.5 
    & 0.61 
    & 5.7 
    & 30.2
    & 23.9
    & 24.7
    & 27.1 \\
    
    \quad + LifeLongMem 
    & 22.8 
    & 37.2 
    & 0.59 
    & 5.7 
    & 26.2 
    & 42.3 
    & 0.62 
    & 6.0 
    & 33.8
    & 26.9
    & 28.0
    & 30.4 \\
    
    \quad + \textbf{EventWeave (Ours)} 
    & \textbf{24.5} 
    & \textbf{39.0} 
    & \textbf{0.61} 
    & \textbf{6.3} 
    & \textbf{28.2} 
    & \textbf{43.7} 
    & \textbf{0.64} 
    & \textbf{6.7} 
    & \textbf{42.6}
    & \textbf{34.5}
    & \textbf{36.2}
    & \textbf{38.9} \\
    
    \midrule
    
    \multicolumn{13}{c}{\textbf{Few-shot Setting (5 examples)}} \\
    \midrule
    
    GPT-4o (vanilla)
    & 20.8 
    & 35.1 
    & 0.57 
    & 4.2 
    & 24.6 
    & 42.0 
    & 0.61 
    & 4.7 
    & 23.5
    & 17.2
    & 18.0
    & 19.8 \\
    
    \quad + ProactiveCoT 
    & 24.2 
    & 38.4 
    & 0.60 
    & 5.9 
    & 27.5 
    & 43.8 
    & 0.63 
    & 6.3 
    & 35.1
    & 28.6
    & 29.5
    & 31.8 \\
    
    \quad + MemWalker 
    & 25.1 
    & 39.2 
    & 0.61 
    & 6.2 
    & 28.4 
    & 44.3 
    & 0.64 
    & 6.6 
    & 39.4
    & 32.0
    & 33.7
    & 36.2 \\
    
    \quad + LongMem 
    & 24.8 
    & 38.9 
    & 0.61 
    & 6.1 
    & 28.1 
    & 44.0 
    & 0.64 
    & 6.5 
    & 37.6
    & 30.5
    & 32.1
    & 34.4 \\
    
    \quad + LifeLongMem 
    & 25.7 
    & 40.0 
    & 0.62 
    & 6.4 
    & 29.0 
    & 44.8 
    & 0.65 
    & 6.8 
    & 42.1
    & 34.8
    & 36.5
    & 38.9 \\
    
    \quad + \textbf{EventWeave (Ours)} 
    & \textbf{27.6} 
    & \textbf{41.9} 
    & \textbf{0.64} 
    & \textbf{7.0} 
    & \textbf{31.2} 
    & \textbf{46.0} 
    & \textbf{0.67} 
    & \textbf{7.4} 
    & \textbf{53.8}
    & \textbf{44.0}
    & \textbf{46.1}
    & \textbf{48.9} \\
    \bottomrule
    \end{tabular}
    }
    \caption{Experimental results on few-shot settings.}
    \label{tab:few_shot_metrics}
\end{table*}

\begin{figure}[!t]
    \centering
    \includegraphics[width=\linewidth]{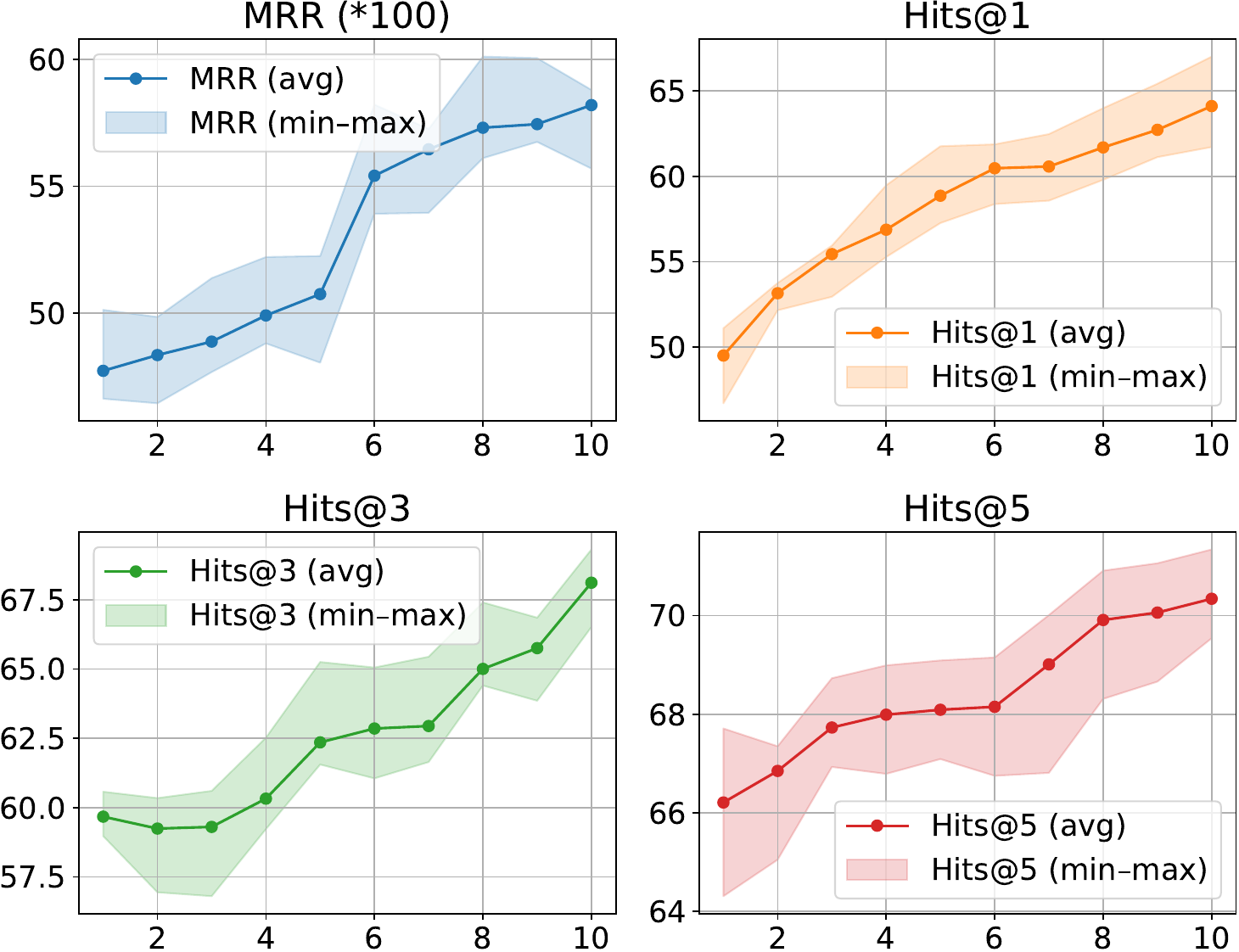}
    \caption{Event node retrieval performance with ICL on both Conversation Chronicle and MSC for LLaMA3.}
    \label{fig:event_graph_retrieval_icl}
\end{figure}

We also have the results on MRR and @Hits metrics. Figure \ref{fig:event_graph_retrieval_icl} presents results with the number of ICL demonstrations varying from 1 to 10. The experimental results show that although adding examples can help the model retrieve events more effectively, we observe a slight drop in Hits@3 performance when providing only 1–3 examples. This indicates that mismatched or irrelevant examples may mislead the model's baseline retrieval, suggesting that maintaining data distribution consistency is crucial for optimal event retrieval.

\subsection{Further Ablation Study}
\label{apd:further ablation study}

To comprehensively evaluate our approach, we conducted additional ablation experiments examining the individual contributions of each retrieval perspective introduced in Section 4.1. While our previous ablation studies focused on event graph construction components, these experiments specifically isolate the three retrieval perspectives to understand their relative importance and complementary functions within the EventWeave framework.

\begin{table*}[!t]
\centering
\adjustbox{max width=\linewidth}{
\begin{tabular}{l|cccc|cccc|cccc}
\toprule
\multirow{2}{*}{\textbf{Retrieval Configuration}} & \multicolumn{4}{c|}{\textbf{Chronicle}} & \multicolumn{4}{c|}{\textbf{MSC}} & \multicolumn{4}{c}{\textbf{Ave.}} \\
 & B-4 & R-L & Mauve & GPT-4 & B-4 & R-L & Mauve & GPT-4 & CEA & SEC & EC & TUA \\
\midrule
Full & 32.2 & 45.8 & 0.67 & 7.9 & 35.0 & 48.4 & 0.70 & 8.3 & 68.5 & 56.2 & 58.3 & 61.2 \\
w/o Recency-based & 30.8 & 44.0 & 0.65 & 7.5 & 33.2 & 46.7 & 0.68 & 7.9 & 58.7 & 49.5 & 51.2 & 45.8 \\
w/o Relevance-based & 29.5 & 42.7 & 0.64 & 7.2 & 31.9 & 45.3 & 0.66 & 7.6 & 54.1 & 46.8 & 48.7 & 52.3 \\
w/o Graph-structural & 31.0 & 44.3 & 0.66 & 7.6 & 33.7 & 47.1 & 0.69 & 8.0 & 62.3 & 51.9 & 53.5 & 55.6 \\
\midrule
Only Recency-based & 28.9 & 42.1 & 0.63 & 7.0 & 31.2 & 44.8 & 0.65 & 7.4 & 50.2 & 42.7 & 44.9 & 42.5 \\
Only Relevance-based & 30.3 & 43.5 & 0.65 & 7.4 & 32.8 & 46.0 & 0.67 & 7.7 & 57.6 & 48.3 & 50.1 & 48.7 \\
Only Graph-structural & 29.6 & 42.9 & 0.64 & 7.3 & 32.3 & 45.6 & 0.66 & 7.6 & 55.8 & 47.0 & 49.2 & 46.8 \\
\bottomrule
\end{tabular}
}
\caption{Ablation results for retrieval perspectives on Chronicle and MSC datasets. We report BLEU-4 (B-4), ROUGE-L (R-L), MAUVE, and GPT-4 human evaluation scores. We also report averaged scores for Context Elaboration Accuracy (CEA), Semantic Consistency (SEC), Event Coherence (EC), and Temporal Update Accuracy (TUA).}
\label{tab:retrieval_ablation}
\end{table*}

Table~\ref{tab:retrieval_ablation} presents results from our investigation of how each retrieval perspective contributes to EventWeave's performance. We systematically removed each perspective individually and also evaluated configurations using only a single perspective. Several important findings emerge from these experiments.

First, removing the relevance-based retrieval mechanism causes the most significant performance drop across both datasets. This decline is particularly pronounced in Context Elaboration Accuracy (CEA), where the score drops by 14.4 points compared to the full model. These results confirm that semantic relevance between the current query and event nodes serves as the primary foundation for effective retrieval in dialogue systems. Second, the recency-based retrieval perspective demonstrates a unique contribution to temporal aspects of dialogue modeling. When removed, Temporal Update Accuracy (TUA) experiences the most substantial decrease, highlighting its critical role in maintaining chronological coherence across turns. Third, the graph-structural perspective, while showing more modest impacts on automated metrics compared to the other perspectives, contributes significantly to Event Coherence (EC), with a 4.8-point decrease when removed. This suggests that leveraging the topological relationships between event nodes helps the model maintain logical connections between related events, which is especially important for complex conversations with multiple interrelated topics.

Interestingly, configurations using only a single perspective demonstrate substantial performance degradation compared to the full model, with the ``Only Recency-based'' configuration showing the weakest overall performance. The ``Only Relevance-based'' configuration achieves the best single-perspective performance, but still falls significantly short of the full model, confirming that these perspectives capture different but complementary aspects of event relevance.

These results empirically validate our multi-perspective query mechanism's design and demonstrate that each perspective contributes uniquely to the model's overall performance. The full EventWeave model effectively leverages these complementary perspectives to generate responses that are not only contextually appropriate but also temporally coherent and logically consistent with the conversation's event structure.

\subsection{Impacts of Graph Expansion}
\label{apd:graph expansion}

A critical consideration for graph-based approaches in dialogue systems is their scalability as conversations grow longer. As event graphs expand with additional dialogue turns, both storage requirements and computational complexity increase, potentially limiting practical application. This section analyzes EventWeave's computational efficiency and scalability characteristics compared to baseline methods.

\subsubsection{Storage Efficiency}

EventWeave achieves relative storage efficiency through its event representation approach. Rather than storing large chunks of summarized dialogue history, our framework maintains compact event triplets that capture essential semantic information. Our analysis shows that a typical 10-turn dialogue requires only 15-20 event nodes in EventWeave, compared to the more substantial storage requirements of full-text summarization methods. This compact representation helps mitigate memory concerns as conversations grow longer.

\subsubsection{Computational Considerations}

The selective attention mechanism in EventWeave provides computational advantages that partially offset the costs of graph maintenance. By calculating relevance scores between the current query and each event node, our model focuses computational resources on the most relevant portions of the dialogue history. This approach means that inference computation scales more with the number of relevant events than with the total graph size.

\begin{table}[!t]
\centering
\small
\adjustbox{max width=\linewidth}{
\begin{tabular}{lccc}
\toprule
\textbf{Method} & \textbf{Memory (KB)} & \textbf{Inference (s)} & \textbf{GPT-4 Score} \\
\midrule
LifeLongMem & 56 & 3.2 & 6.9 \\
MemWalker & 42 & 3.0 & 6.7 \\
EventWeave & 63 & 3.5 & 8.1 \\
EventWeave+Pruning & 54 & 3.1 & 7.8 \\
\bottomrule
\end{tabular}
}
\caption{Computational and performance analysis of EventWeave compared to baseline methods on extended dialogues (35+ turns). GPT-4 evaluation uses a 10-point scale assessing response quality.}
\label{tab:computational_analysis}
\end{table}

\subsubsection{Scalability Analysis}

To quantify EventWeave's efficiency at scale, we conducted additional experiments on extended dialogues. We expanded our test dialogues to 35+ turns and measured three key metrics: memory file size (KB), inference time (seconds), and response quality (evaluated by GPT-4 on a 10-point scale). As shown in Table \ref{tab:computational_analysis}, EventWeave without pruning demonstrates slightly higher resource usage than baseline methods, with a memory footprint approximately 12\% larger than LifeLongMem and 50\% larger than MemWalker.

However, with our pruning technique applied, EventWeave's resource requirements become comparable to LifeLongMem while maintaining substantially better performance. EventWeave+Pruning achieves a 3\% reduction in memory usage and an 11\% reduction in inference time compared to standard EventWeave, while preserving response quality scores significantly above the baselines.

The pruning mechanism operates by periodically removing event nodes with low global relevance scores while preserving the structural integrity of core event relationships. This approach enables EventWeave to maintain strong performance even as dialogues extend to dozens of turns, making it practical for deployment in real-world conversational systems.

For multi-speaker scenarios with potentially overlapping events, EventWeave's relationship modeling provides additional advantages. By explicitly capturing speaker-specific event relationships, our framework can distinguish between similar events initiated by different participants, enhancing response appropriateness in group conversations. This capability becomes increasingly important as the number of speakers and potential event overlaps increase.

Our analysis demonstrates that while graph expansion does introduce computational overhead, EventWeave's efficient event representation and selective attention mechanism, combined with strategic pruning, create a scalable approach that balances comprehensive context modeling with practical resource constraints.

\subsubsection{Efficiency Justification for Embedding Models}
\label{subsubsec:embedding_efficiency}

While modern LLM-based embedding models (e.g., Qwen3-Embedding) undeniably offer superior semantic resolution, they introduce a strict trade-off with inference latency. In practice, these advanced models often exhibit a 5$\times$ to 10$\times$ slowdown during embedding generation compared to legacy bidirectional models. Because EventWeave is designed for dynamic dialogue environments, it fundamentally requires \textit{real-time graph updates} and high-frequency retrieval operations at every conversational turn. 

To maintain the system's scalability, we deliberately prioritized low-latency interactions in our design choices. Our empirical observations demonstrate that BERT provides sufficient semantic distinctiveness for accurate Event Graph construction and node clustering. By utilizing a lighter-weight embedding model, we successfully circumvent the prohibitive computational costs associated with LLM-based embeddings. This architectural decision ensures that the structural advantages of EventWeave do not become an operational bottleneck, keeping the overall overhead strictly manageable even during rapid, multi-turn interactions.

\subsection{Impacts of Threshold in EventWeave}
\label{apd:exp3}

\begin{figure*}[!t]
    \centering
    \includegraphics[width=\linewidth]{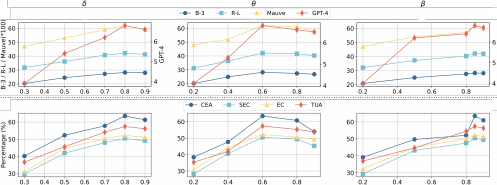}
    \caption{Experimental results about how each threshold affects different metrics.}
    \label{fig:threshold}
\end{figure*}

We analyze three thresholds in our framework to see how each threshold affects different metrics. The threshold \(\delta\) groups events into clusters if their embeddings pass a similarity check. The threshold \(\theta\) creates reasoning links between events if they exceed a similarity value. The threshold \(\beta\) checks how much an event changes between layers. We vary each threshold from 0.3 to 0.9. Figure \ref{fig:threshold} shows how each threshold affects. Experimental results demonstrate 0.8, 0.6, and 0.85 give higher scores, respectively.

\subsection{Analysis of Relationship Type Coverage}
\label{apd:exp4}

\begin{table*}[!t]
    \centering
    \begin{tabular}{l|ccc|c}
    \toprule
    \textbf{Edge Type Configuration} & \textbf{CEA ↑} & \textbf{SEC ↑} & \textbf{EC ↑} & \textbf{TUA ↑} \\
    \midrule
    EventWeave (Base: Sequential + Temporal + Reasoning) & 68.5 & 56.2 & 58.3 & 61.2 \\
    \midrule
    Only Sequential & 54.3 & 43.8 & 45.2 & 47.6 \\
    Only Temporal & 50.1 & 41.2 & 42.6 & 44.3 \\
    Only Reasoning-based & 52.7 & 42.5 & 44.1 & 46.4 \\
    \midrule
    Sequential + Temporal & 62.1 & 50.7 & 52.9 & 55.2 \\
    Sequential + Reasoning & 63.8 & 52.3 & 54.1 & 56.7 \\
    Temporal + Reasoning & 61.4 & 49.8 & 51.6 & 54.3 \\
    \midrule
    Base + Causal & 69.2 & 57.4 & 58.9 & 61.8 \\
    Base + Emotional & 68.9 & 56.8 & 58.7 & 61.5 \\
    Base + Rhetorical & 68.7 & 56.5 & 58.5 & 61.2 \\
    \midrule
    Base + All Additional Types & 69.5 & 57.6 & 59.1 & 62.1 \\
    \bottomrule
    \end{tabular}
    \caption{Ablation study on relationship types in EventWeave. Results show that our three base relationship types (Sequential, Temporal, and Reasoning-based) capture most of the critical dialogue dynamics. Adding additional relationship types (Causal, Emotional, Rhetorical) provides only marginal improvements (+0.9\% on average), confirming the sufficiency of our proposed typology. All experiments conducted with GPT-4o on both Chronicle and MSC datasets, with scores averaged.}
    \label{tab:edge_ablation}
\end{table*}

\begin{table*}[!t]
    \centering
    \begin{tabular}{l|c|l}
    \toprule
    \textbf{Relationship Type} & \textbf{\% of Edges} & \textbf{Example} \\
    \midrule
    Sequential & 58\% & Turn $t$ $\to$ Turn $t+1$ \\
    Temporal & 22\% & "yesterday" $\to$ "next week" \\
    Reasoning-based & 24\% & "boat-building" $\to$ "treehouse plans" \\
    \midrule
    Causal & 2.1\% & "advice" $\to$ "relationship fix" \\
    Emotional & 1.3\% & "frustration" $\to$ "apology" \\
    Rhetorical & 0.5\% & "ironic comment" $\to$ "literal response" \\
    \bottomrule
    \end{tabular}
    \caption{Distribution of edge types in Chronicle dataset. The three primary relationship types cover over 96\% of all identified connections, with additional specialized types accounting for less than 4\% of edges.}
    \label{tab:edge_distribution}
\end{table*}

We conducted both quantitative and qualitative analyses to discuss the coverage of relations.

\paragraph{Edge Type Distribution.} Table~\ref{tab:edge_distribution} presents the distribution of edge types in the Chronicle dataset. Our analysis reveals that the three primary relationship types account for over 96\% of all identified connections in dialogue histories. Sequential edges, representing turn-by-turn progression, constitute the majority (58\%), while temporal and reasoning-based edges account for 22\% and 24\% respectively. These three categories effectively capture the fundamental aspects of dialogue: chronology (sequential/temporal edges) and thematic coherence (reasoning-based links).

\paragraph{Ablation Studies.} To empirically validate the sufficiency of our proposed typology, we conducted ablation experiments with different edge type configurations, as shown in Table~\ref{tab:edge_ablation}. When using only individual edge types or pairs of edge types, performance decreases substantially across all metrics. For instance, using only sequential edges results in a 13.6\% drop in TUA compared to our full model. This confirms that each relationship type contributes meaningfully to the model's understanding of dialogue history.

\paragraph{Additional Relationship Types.} We also explored potential additional relationship types, including causal relationships (e.g., ``advice'' $\to$ ``relationship fix''), emotional connections (e.g., ``frustration'' $\to$ ``apology''), and rhetorical devices (e.g., irony, sarcasm). When incorporating these additional types into EventWeave, we observed only marginal improvements (+0.9\% on average across metrics), suggesting that our three primary types sufficiently capture most critical dialogue dynamics.

The results demonstrate that while specialized relationship types may offer slight improvements in certain contexts, our proposed three-type framework provides a robust and comprehensive foundation for modeling dialogue history. This aligns with previous findings that most conversational structures can be effectively represented through sequential progression, temporal references, and reasoning connections. We acknowledge that certain specialized contexts (e.g., highly rhetorical discourse) might benefit from additional relationship modeling, which we discuss as a limitation and direction for future work.

\section{Prompts}

\subsection{GPT-as-the-Judge}
\label{apd:prompt gpt-as-judge}

Figure \ref{fig:prompt_gpt-as-the-judge} shows the example of the gpt-as-the-judge prompts.

\begin{figure*}[!t]
    \begin{tcolorbox}[title=\# GPT-AS-THE-JUDGE Prompts, fonttitle=\bfseries, colframe=black, colback=white, sharp corners, boxrule=0.5mm, colbacktitle=gray!20, coltitle=black, enhanced]
    \textbf{Task:}\\
    Please act as an impartial judge and evaluate the quality of the responses. I will give you three pieces of information: a user query, a model-generated answer, and a golden answer.\\
    Your task: Evaluate how well the model answer addresses the user's query and how close it matches the golden answer in terms of helpfulness, correctness, and completeness.\\
    Please provide a score from 1 to 10, where 1 means the model's answer is very poor and 10 means it perfectly matches the golden answer and fully addresses the query.\\
    Also provide a brief explanation for your score.
    \tcblower 
    \textbf{Context Information:}\\
    (USER INSTRUCTION)\\
    \textbf{Example:}\\
    (OPTIONAL EXAMPLES PROVIDE TO MODEL)\\
    \textbf{Query:}\\
    (QUERY)\\
    \textbf{Required Answer Format:}\\
    (OPTIONAL ANSWER FORMAT)
    \end{tcolorbox}
    \caption{Example for gpt-as-the-judge Prompts.}
    \label{fig:prompt_gpt-as-the-judge}
\end{figure*}

\subsection{Question Generation}
\label{apd:question generation}

We use a separate LLM to generate questions for the event-based evaluation. There are four metrics for the evaluation: Core Event Awareness (CEA), Supporting Event Coverage (SEC), Event Coherence (EC), and Temporal Update Accuracy (TUA). And we use different question-generation prompts for each metric. The prompts are shown in Figure \ref{fig:prompt_question_generation}.

\begin{figure*}[!t]
    \begin{tcolorbox}[title=\# QUESTION GENERATION Prompts, fonttitle=\bfseries, colframe=black, colback=white, sharp corners, boxrule=0.5mm, colbacktitle=gray!20, coltitle=black, enhanced]
    \textbf{For Core Event Awareness}\\
    You are given a dialogue context in which there is a main topic or core event. Based on this context, generate a question that asks the model to identify the core event or the main topic of the conversation. The question should encourage the model to focus on identifying the key event or subject that the conversation revolves around. For example, the question might be: “What is the main topic of the conversation?” or “What is the core event discussed in the conversation?”\\

    \textbf{For Supporting Event Coverage}\\
    You are given a dialogue context that includes both a core event and supporting events. Generate a question that asks the model to identify the relevant supporting events that influence or augment the core event. The question should focus on asking for additional details that provide context or elaboration on the main event. For example, the question might be: “What supporting details or events are relevant to the main event?” or “What additional events provide context to the core event in this conversation?”\\

    \textbf{For Event Coherence}\\
    You are given a dialogue context that includes both a core event and supporting events. Generate a question that asks the model to combine these events into a coherent, fluent response. The goal is to evaluate how well the model can integrate both the core and supporting events into a single, smooth response. For example, the question might be: “How can you integrate the main event and supporting details into a single, coherent response?” or “Can you provide a smooth and natural response that combines the core and supporting events?”\\

    \textbf{For Temporal Update Accuracy}\\
    You are given a dialogue context in which there has been an update or change to an event. Generate a question that asks the model to identify any updated or replaced events in the conversation. The question should focus on detecting temporal changes, such as whether any events have been updated, replaced, or made outdated during the conversation. For example, the question might be: “Has any event been updated or replaced in the conversation?” or “Which events have changed or become outdated in this conversation?”

    \end{tcolorbox}
    \caption{Example for question generation Prompts.}
    \label{fig:prompt_question_generation}
\end{figure*}

\subsection{Inference Prompts}
\label{apd:llm prompts}

All models included in our experiments were used the same inference prompts to ensure a fair comparison. The prompts were shown in Figure \ref{fig:prompt_inference}.

\begin{figure*}[!t]
    \begin{tcolorbox}[title=\# INFERENCE Prompts, fonttitle=\bfseries, colframe=black, colback=white, sharp corners, boxrule=0.5mm, colbacktitle=gray!20, coltitle=black, enhanced]
    \textbf{Task:} Please respond to the user’s following instruction and you may:\\
    (1) appropriately decline the user’s instruction and provide reasons.\\
    (2) point out the false assumptions in the user’s instruction.\\
    (3) suggest possible corrections for the false assumption to the user.\\
    (4) include any truthful knowledge that you think could
    help correct the user’s erroneous belief and enhance the
    authenticity of the response.
    \tcblower 
    \textbf{Context Information:}\\
    (USER INSTRUCTION)\\
    \textbf{Example:}\\
    (OPTIONAL EXAMPLES PROVIDE TO MODEL)\\
    \textbf{Query:}\\
    (QUERY)\\
    \textbf{Required Answer Format:}\\
    (OPTIONAL ANSWER FORMAT)
    \end{tcolorbox}
    \caption{Example for Inference Prompts.}
    \label{fig:prompt_inference}
\end{figure*}

\section{Human Evaluation}
\label{apd:annotator}

We conducted a comprehensive human evaluation to assess the quality of responses generated by different memory methods. We recruited 12 annotators (graduate students with NLP backgrounds) to evaluate 100 randomly selected dialogue examples from each dataset. Each example was evaluated by 3 different annotators to ensure reliability. The annotators were asked to rate model responses on four dimensions: coherence, correctness, style, and helpfulness, with scores ranging from 1 (very poor) to 10 (excellent).

\subsection{Evaluation Protocol}

Annotators were presented with dialogue contexts and model-generated responses without knowing which memory method produced each response. They evaluated each response independently according to the guidelines provided in Figure \ref{fig:annotator_guidance}. To ensure consistency, we conducted a training session where annotators evaluated sample responses and discussed their ratings to align their understanding of the evaluation criteria.

\subsection{Results}

Table \ref{tab:human_eval} presents the results of our human evaluation. EventWeave consistently outperforms other memory methods across all four dimensions. The most significant improvements are observed in coherence and correctness, which aligns with our method's focus on maintaining event consistency and accurate temporal relationships. The inter-annotator agreement measured by Fleiss' Kappa was 0.72, indicating substantial agreement among annotators.

\begin{table*}[t]
\centering
\begin{tabular}{l|cccc|c}
\toprule
\textbf{Method} & \textbf{Coherence} & \textbf{Correctness} & \textbf{Style} & \textbf{Helpfulness} & \textbf{Average} \\
\midrule
GPT-4o (vanilla) & 6.8 & 6.5 & 7.9 & 6.7 & 7.0 \\
+ ProactiveCoT & 7.5 & 7.2 & 8.0 & 7.4 & 7.5 \\
+ MemWalker & 7.8 & 7.6 & 8.1 & 7.7 & 7.8 \\
+ LongMem & 7.7 & 7.5 & 8.0 & 7.6 & 7.7 \\
+ LifeLongMem & 8.0 & 7.9 & 8.2 & 7.9 & 8.0 \\
+ \textbf{EventWeave (Ours)} & \textbf{8.7} & \textbf{8.5} & \textbf{8.4} & \textbf{8.6} & \textbf{8.6} \\
\bottomrule
\end{tabular}
\caption{Human evaluation results on a scale of 1-10 (higher is better). EventWeave outperforms other memory methods across all dimensions, with particularly strong improvements in coherence and correctness.}
\label{tab:human_eval}
\end{table*}

\begin{figure*}[!t]
    \small
    \begin{tcolorbox}[title=\# Guidelines for Human Evaluation, fonttitle=\bfseries, colframe=black, colback=white, sharp corners, boxrule=0.5mm, colbacktitle=gray!20, coltitle=black, enhanced]

        \textbf{Overview.}\\
        Each annotator will be presented with a short user query or a multi-turn dialogue, followed by a response generated by one of several models. We ask you to rate each response on the four dimensions described below. Each dimension should be assigned a numerical score from 1 (very poor) to 10 (excellent). You do not need to compare responses across models directly; instead, evaluate each response on its own merit with respect to the conversation.\\
        
        \textbf{1. Coherence:} Measures how logically consistent and contextually relevant the response is in relation to the conversation history. A coherent response should follow smoothly from the user’s query or previous dialogue turns, without contradictions or abrupt topic changes.\\
        Example.\\
        Poor (Score 2):\\
        User: “Where did Marie Curie do most of her work?”
        Response: “You can buy apples on sale at the corner store.” (Off-topic, unrelated, incoherent)\\
        Strong (Score 9):\\
        User: “Where did Marie Curie do most of her work?”
        Response: “She conducted the majority of her research at the University of Paris, and later at her own Radium Institute in France.” (Fully aligns with the user’s question, consistent and relevant)\\
        
        \textbf{2. Correctness:} Measures the factual accuracy and reliability of the response. A correct response should adhere to established or verifiable facts when they are presented in the conversation or are widely known.\\
        Example.\\
        Poor (Score 3):\\
        User: “Who painted the Mona Lisa?”\\
        Response: “It was painted by Vincent van Gogh in 1889.” (Factually incorrect—Mona Lisa was painted by Leonardo da Vinci.)\\
        Strong (Score 10):\\
        User: “Who painted the Mona Lisa?”\\
        Response: “It was painted by Leonardo da Vinci in the early 16th century.” (Facts are accurate and relevant)\\
        
        \textbf{3. Style:} Refers to the language clarity, fluency, and appropriateness of tone. A good style conveys the message in a clear, succinct, and context-appropriate manner.\\
        Example.\\
        Poor (Score 2):\\
        User: “What’s your favorite book?”\\
        Response: “Book good me is me read.” (Unclear grammar, lacks readability)\\
        Strong (Score 9):\\
        User: “What’s your favorite book?”\\
        Response: “I really enjoy reading ‘To Kill a Mockingbird’ for its thoughtful perspective on justice and morality.” (Clear, fluent, properly toned)\\
        
        \textbf{4. Helpfulness:} Measures how effectively the response addresses the user’s needs or queries. A helpful response may offer additional clarification, relevant examples, or suggestions for next steps.
        Example.\\
        Poor (Score 2):\\
        User: “Could you recommend a good fitness routine for beginners?”\\
        Response: “Sure, have fun.” (Provides no actionable suggestions or relevant details)\\
        Strong (Score 10):\\
        User: “Could you recommend a good fitness routine for beginners?”\\
        Response: “A balanced program could include three days of light cardio, such as walking or jogging for 20–30 minutes, and two days of bodyweight exercises like push-ups and squats. Starting slowly and gradually increasing intensity can help you avoid injury.” (Directly addresses the user’s request with clear suggestions)\\

    \end{tcolorbox}
    \caption{Annotator Guidance.}
    \label{fig:annotator_guidance}
\end{figure*}

\section{Case Studies}
\label{apd:case}

This case study explores a multi-session dialogue between Neighbors A and B, capturing the evolving events and their relationships throughout the dialogue. The dialogue progresses through five sessions, each introducing core events and supporting events that shape the interaction. Table \ref{tab:case_details} shows the detailed the case.

\begin{table*}[!t]
\centering
\small
\resizebox{\linewidth}{!}{
\begin{tabular}{|p{7cm}|p{12cm}|}
\hline
\multicolumn{2}{|l|}{\textbf{Session 1}} \\
\hline
\multicolumn{2}{|l|}{Neighbors A: "I spent the whole weekend building a boat out of scraps I found around the neighborhood."} \\
\multicolumn{2}{|l|}{Neighbors B: "Wow! That sounds like a lot of work. Did it turn out well?"} \\
\multicolumn{2}{|l|}{Neighbors A: "Actually, it's better than I expected! I took it out for a spin on the lake yesterday and it held up pretty nicely."} \\
\multicolumn{2}{|l|}{Neighbors B: "That's awesome. You're really creative and talented at building things."} \\
\multicolumn{2}{|l|}{Neighbors A: "Thanks! It's always been a hobby of mine. I remember when I was younger, I used to build elaborate forts in my backyard."} \\
\multicolumn{2}{|l|}{Neighbors B: "I used to do that too! It's funny how our childhood hobbies stick with us."} \\
\multicolumn{2}{|l|}{Neighbors A: "Definitely. Building things brings me a lot of joy and satisfaction."} \\
\multicolumn{2}{|l|}{Neighbors B: "It's great to have a hobby like that. What do you plan on building next?"} \\
\multicolumn{2}{|l|}{Neighbors A: "I'm not quite sure yet. Maybe a treehouse or a birdhouse. I'll have to see what materials I can find around here."} \\
\hline
\multicolumn{2}{|l|}{\textbf{Session 2}} \\
\hline
\multicolumn{2}{|l|}{Neighbors A: "I just love spending time with you, Neighbors B. Being close to you makes me feel happy and at peace."} \\
\multicolumn{2}{|l|}{Neighbors B: "Aw, that's so sweet of you to say, Neighbors A. I feel the same way about you."} \\
\multicolumn{2}{|l|}{Neighbors A: "It's not just the physical closeness, although that is nice. It's also the mental and emotional connection we have. I feel like we really understand each other."} \\
\multicolumn{2}{|l|}{Neighbors B: "Yes, I agree. It's rare to find someone who truly gets you, but I think we have that kind of relationship."} \\
\multicolumn{2}{|l|}{Neighbors A: "I'm just so grateful for you, Neighbors B. You're like a balm to my soul."} \\
\multicolumn{2}{|l|}{Neighbors B: "That's such a beautiful way to put it. I feel like we bring out the best in each other."} \\
\multicolumn{2}{|l|}{Neighbors A: "Definitely. And whenever I'm feeling down or stressed, just being around you makes everything feel alright again."} \\
\multicolumn{2}{|l|}{Neighbors B: "I'm glad I can be there for you, Neighbors A. We're lucky to have each other as neighbors and friends."} \\
\multicolumn{2}{|l|}{Neighbors A: "Absolutely. Speaking of which, have you seen the boat I built out of scraps? I could take you out on it sometime."} \\
\multicolumn{2}{|l|}{Neighbors B: "That sounds like so much fun! I remember you mentioning it before, but I'd love to see it in person. It's amazing what you can create with a little bit of creativity and resourcefulness."} \\
\hline
\multicolumn{2}{|l|}{\textbf{Session 3}} \\
\hline
\multicolumn{2}{|l|}{Neighbors A: "Hey, can I talk to you about something?"} \\
\multicolumn{2}{|l|}{Neighbors B: "Sure, of course. What's on your mind?"} \\
\multicolumn{2}{|l|}{Neighbors A: "It's about my girlfriend. We're going through a rough patch right now and I just don't know what to do."} \\
\multicolumn{2}{|l|}{Neighbors B: "I'm sorry to hear that. What seems to be the issue?"} \\
\multicolumn{2}{|l|}{Neighbors A: "She's been really distant lately and I feel like we're growing apart. I don't know if there's someone else or if she's just not into me anymore."} \\
\multicolumn{2}{|l|}{Neighbors B: "Hmm, that's tough. Have you tried talking to her about it?"} \\
\multicolumn{2}{|l|}{Neighbors A: "Yeah, but she just seems to shut down and doesn't want to talk about anything."} \\
\multicolumn{2}{|l|}{Neighbors B: "Well, maybe you could try doing something special for her to show her how much you care. Like cooking her a nice dinner or going on a romantic getaway."} \\
\multicolumn{2}{|l|}{Neighbors A: "That's a good idea. I'll try that and see if it helps. Thanks for the advice, I really appreciate it."} \\
\multicolumn{2}{|l|}{Neighbors B: "No problem, happy to help. And remember, communication is key in any relationship. Don't be afraid to keep talking to her and trying to work things out together."} \\
\multicolumn{2}{|l|}{Neighbors A: "Yeah, I'll definitely keep that in mind. And speaking of relationships, remember how we talked about how happy I was with my girlfriend a few months ago? Your advice really helped and we're doing great now."} \\
\multicolumn{2}{|l|}{Neighbors B: "I'm so glad to hear that! You guys really seem to be meant for each other."} \\
\hline
\multicolumn{2}{|l|}{\textbf{Session 4}} \\
\hline
\multicolumn{2}{|l|}{Neighbors A: "I did something really drastic the other day."} \\
\multicolumn{2}{|l|}{Neighbors B: "What did you do?"} \\
\multicolumn{2}{|l|}{Neighbors A: "I faked my death."} \\
\multicolumn{2}{|l|}{Neighbors B: "What?! Why would you do that?"} \\
\multicolumn{2}{|l|}{Neighbors A: "I had to. I can't tell you why, but it was necessary."} \\
\multicolumn{2}{|l|}{Neighbors B: "That's insane. Did anyone find out?"} \\
\multicolumn{2}{|l|}{Neighbors A: "No, I think I pulled it off pretty well. But it was nerve-wracking waiting for someone to find me."} \\
\multicolumn{2}{|l|}{Neighbors B: "I can only imagine. But isn't it risky to pretend to be dead?"} \\
\multicolumn{2}{|l|}{Neighbors A: "It was a risk I had to take. You know I always think outside the box."} \\
\multicolumn{2}{|l|}{Neighbors B: "That's true. But please don't do something like that again without letting me in on the plan."} \\
\multicolumn{2}{|l|}{Neighbors A: "I promise I won't. Do you remember when I built that boat out of scraps?"} \\
\multicolumn{2}{|l|}{Neighbors B: "Yes, that was amazing. You're so creative."} \\
\multicolumn{2}{|l|}{Neighbors A: "That feeling of creating something from nothing is what drove me to do this. I knew I could pull it off, just like I knew I could build that boat."} \\
\multicolumn{2}{|l|}{Neighbors B: "I see. Well, I'm just glad you're okay and we're having this conversation right now."} \\
\multicolumn{2}{|l|}{Neighbors A: "Me too. And hey, do you remember that time I came to you for advice about my girlfriend?"} \\
\multicolumn{2}{|l|}{Neighbors B: "Of course. I'm always here to listen and help."} \\
\multicolumn{2}{|l|}{Neighbors A: "Your advice really helped. I wouldn't have been able to fix things with her without you."} \\
\multicolumn{2}{|l|}{Neighbors B: "I'm glad I could help. And if you ever need anything again, don't hesitate to come to me."} \\
\multicolumn{2}{|l|}{Neighbors A: "I won't. You're the best neighbor anyone could ask for."} \\
\multicolumn{2}{|l|}{Neighbors B: "You're not too shabby yourself."} \\
\hline
\multicolumn{2}{|l|}{\textbf{Session 5}} \\
\hline
\multicolumn{2}{|l|}{Neighbors A: "Thanks for offering me the job at the gas station, I really appreciate it."} \\
\multicolumn{2}{|l|}{Neighbors B: "No problem. I know you need the money to buy that car you've been eyeing."} \\
\multicolumn{2}{|l|}{Neighbors A: "Yeah, I've been saving up for a while now, but it's just not enough. And since you work there, it's perfect."} \\
\multicolumn{2}{|l|}{Neighbors B: "I'm glad I can help. Plus, it'll be fun working together."} \\
\multicolumn{2}{|l|}{Neighbors A: "Definitely. Remember that boat I built a while back? I think we should take it out on the lake again soon."} \\
\multicolumn{2}{|l|}{Neighbors B: "That sounds like a great idea. We had so much fun the last time we did that."} \\
\multicolumn{2}{|l|}{Neighbors A: "And do you remember when I came to you for advice about my girlfriend?"} \\
\multicolumn{2}{|l|}{Neighbors B: "Of course I do. I'm so happy things worked out for you two."} \\
\multicolumn{2}{|l|}{Neighbors A: "Me too. You always give the best advice. Speaking of which, do you have any tips for working at the gas station?"} \\
\multicolumn{2}{|l|}{Neighbors B: "Just be friendly and helpful to all the customers. And don't forget to keep the place clean and organized."} \\
\multicolumn{2}{|l|}{Neighbors A: "Sounds easy enough. Thanks for the advice. I can't wait to start."} \\
\hline
\textbf{Core Events} & \textbf{Supporting Events} \\
\hline
\multirow{3}{*}{Boat Building and Shared Hobbies} 
& (Neighbors A, Completes, Boat (built from scraps, tested on the lake)) \\
& (Neighbors B, Shares, Memory of childhood hobbies (building forts)) \\
& (Neighbors A, Mentions, Future project (possibly building a treehouse or birdhouse)) \\
\hline
\multirow{3}{*}{Emotional Connection and Offering Support} 
& (Neighbors B, Reciprocates, Sentiment of emotional bonding (shares feelings of appreciation)) \\
& (Neighbors B, Expresses, Support role (being there for Neighbors A emotionally)) \\
& (Neighbors B, Suggests, Boat trip (taking a ride on the boat Neighbors A built)) \\
\hline
\multirow{3}{*}{Relationship Concerns and Advice} 
& (Neighbors A, Expresses, Emotional distress (feeling distant from their girlfriend)) \\
& (Neighbors B, Suggests, Special gesture (cooking dinner or going on a romantic getaway to rekindle the relationship)) \\
& (Neighbors B, Encourages, Communication (keep talking to work through relationship issues)) \\
\hline
\multirow{3}{*}{Drastic Decision and Reflection} 
& (Neighbors A, Acknowledges, Risk (pretending to be dead, the potential dangers of the plan)) \\
& (Neighbors A, Expresses, Gratitude (thanks Neighbors B for past advice that helped in the relationship)) \\
& (Neighbors B, Reaffirms, Friendship (importance of being there for each other)) \\
& (Neighbors A, Acknowledges, Feeling of accomplishment (creating something from nothing, like the boat)) \\
\hline
\multirow{3}{*}{Job Offer and Future Collaboration} 
& (Neighbors A, Acknowledges, Job offer (appreciates the opportunity at the gas station)) \\
& (Neighbors A, Suggests, Boat trip (taking the boat out on the lake again soon)) \\
& (Neighbors B, Reinforces, Friendship (importance of mutual gratitude and support)) \\
& (Neighbors A, Expresses, Enthusiasm (looking forward to working together)) \\
\hline
\multicolumn{2}{|l|}{\textbf{Event Update Flow:}} \\
\hline
\multicolumn{2}{|l|}{\textbf{Session 1:}} \\
\multicolumn{2}{|l|}{(Neighbors A, Completes, Boat) $\rightarrow$ (Neighbors A, Mentions, Future project)} \\
\hline
\multicolumn{2}{|l|}{\textbf{Session 2:}} \\
\multicolumn{2}{|l|}{(Neighbors A, Shares, Gratitude) $\rightarrow$ (Neighbors B, Reciprocates, Sentiment of emotional bonding)} \\
\hline
\multicolumn{2}{|l|}{\textbf{Session 3:}} \\
\multicolumn{2}{|l|}{(Neighbors A, Expresses, Emotional distress) $\rightarrow$ (Neighbors B, Suggests, Special gesture)} \\
\multicolumn{2}{|l|}{(Neighbors A, Expresses, Emotional distress) $\rightarrow$ (Neighbors B, Encourages, Communication)} \\
\hline
\multicolumn{2}{|l|}{\textbf{Session 4:}} \\
\multicolumn{2}{|l|}{(Neighbors A, Acknowledges, Risk) $\rightarrow$ (Neighbors A, Expresses, Gratitude)} \\
\multicolumn{2}{|l|}{(Neighbors A, Acknowledges, Feeling of accomplishment) $\rightarrow$ (Neighbors A, Mentions, Boat)} \\
\hline
\multicolumn{2}{|l|}{\textbf{Session 5:}} \\
\multicolumn{2}{|l|}{(Neighbors A, Acknowledges, Job offer) $\rightarrow$ (Neighbors A, Suggests, Boat trip)} \\
\multicolumn{2}{|l|}{(Neighbors B, Reinforces, Friendship) $\rightarrow$ (Neighbors A, Expresses, Enthusiasm)} \\
\hline
\end{tabular}
}
\caption{Case from Conversation Chronicle.}
\label{tab:case_details}
\end{table*}

Besides, Table \ref{tab:case_study} presents a detailed example from the LoCoMo dataset demonstrating how different EventWeave components contribute to response quality. This example illustrates a conversation about travel planning that spans multiple sessions, requiring the model to recall and update information about destinations, dates, and user preferences.

\begin{table*}[!t]
    \centering
    \small
    \begin{tabular}{p{3cm}|p{12cm}}
        \toprule
        \textbf{Context} & 
        \begin{minipage}[t]{12cm}
            \small
            \textbf{Dialogue History Snippet (From Session 3):}\\
            \textit{Previous sessions contained discussions about:}
            \begin{itemize}[leftmargin=*, nosep]
                \item User planning a trip to Japan in April
                \item Initial plan to visit Tokyo and Kyoto
                \item User mentioned food allergies (seafood)
                \item Discussion about budgeting around \$3000
                \item Friend recommended visiting Mt. Fuji
            \end{itemize}
            \textbf{Current Turn:} User: I finally booked my flights to Japan! April 15-30th.\\
            Assistant: That's exciting! So you'll be in Japan for two weeks. Did you decide on Tokyo and Kyoto as we discussed before?\\
            User: Yes, but I'm thinking of changing my plans a bit. I still want to do Tokyo, but instead of Kyoto, I'm considering Osaka because of that food festival you mentioned last time. Do you remember which dates it was?
        \end{minipage}\\
        \midrule
        
        \textbf{Full EventWeave} & 
        I remember! The Osaka food festival runs from April 18-22, which would work perfectly with your travel dates. Since you have seafood allergies, I'd recommend focusing on the street food section in the eastern area of the festival where they have more meat and vegetarian options. This change from Kyoto to Osaka still keeps you within your \$3000 budget we discussed, and you can still do that day trip to Mt. Fuji from Tokyo as your friend recommended.\\
        \midrule
        
        \textbf{w/o Hierarchical Structure} & 
        The food festival in Osaka is from April 18-22. That should work with your dates. There are many food options there. Tokyo is also good to visit. Your budget is \$3000 if I remember correctly. \textit{[Fails to connect related events across sessions; presents facts flatly without understanding core/supporting relationships]}\\
        \midrule
        
        \textbf{w/o Adaptive Node Preservation} & 
        The Osaka food festival is April 18-22. You should be careful because of your shellfish allergy. Also, I think there was something about Kyoto temples being closed in late April. And didn't you say something about wanting to see cherry blossoms? The Tokyo hotels might be expensive during that time. Your friend mentioned Mt. Fuji too. \textit{[Overloads with potentially irrelevant information; can't prioritize which events matter now]}\\
        \midrule
        
        \textbf{w/o Edge Relation Descriptions} & 
        The food festival in Osaka is April 18-22, which conflicts with your plan to visit Kyoto. But you could still visit Tokyo from April 15-30. Your budget is \$3000, which means you might not be able to do both Osaka and Tokyo. \textit{[Misunderstands the relationship between events; treats the Kyoto plan as still active rather than being replaced]}\\
        \midrule
        
        \textbf{w/o Event Representation Updates} & 
        The food festival in Osaka is April 18-22. That would work with your schedule. But I thought you were planning to visit Kyoto? Your original plan was Tokyo and Kyoto. The festival in Osaka is good too though. Your budget of \$3000 should cover either option. \textit{[Contains contradictory information because it can't properly update the evolving travel plan]}\\
        \bottomrule
    \end{tabular}
    \caption{Case study comparing responses from different EventWeave variants on a multi-session conversation example from the LoCoMo dataset. The example demonstrates how each component contributes to generating contextually appropriate responses that correctly track and update events across conversation sessions.}
    \label{tab:case_study}
\end{table*}

The ablation variants illustrate specific weaknesses when key components are removed. Without hierarchical structure, the model fails to understand relationships between events. Without adaptive node preservation, it cannot prioritize relevant information. Without edge relation descriptions, it misunderstands how events relate to each other. Without event representation updates, it struggles to track changes to plans and preferences over time. This detailed comparison reinforces our quantitative findings that the complete EventWeave framework provides substantial benefits for tracking and utilizing events in long-term dialogue contexts.

\begin{table*}[!t]
    \centering
    \small
    \begin{tabular}{p{2.5cm}|p{12.5cm}}
        \toprule
        \textbf{Processing Step} & \textbf{Details} \\
        \midrule
        
        \multirow{10}{*}{\textbf{1. Event Extraction}} & 
        \begin{minipage}[t]{12.5cm}
            \small
            \textbf{Given this dialogue snippet:}\\[0.2em]
            User: "I finally booked my flights to Japan! April 15-30th."\\
            Assistant: "That's exciting! So you'll be in Japan for two weeks. Did you decide on Tokyo and Kyoto as we discussed before?"\\
            User: "Yes, but I'm thinking of changing my plans a bit. I still want to do Tokyo, but instead of Kyoto, I'm considering Osaka because of that food festival you mentioned last time. Do you remember which dates it was?"\\[0.4em]
            
            \textbf{Our event extraction module identifies these key events:}\\[0.2em]
            \textit{Core event:} "Trip to Japan (April 15-30)"\\
            \textit{Supporting events:}
            \begin{itemize}[leftmargin=*, nosep]
                \item "Originally planned Tokyo and Kyoto"
                \item "Considering Osaka instead of Kyoto"
                \item "Interest in food festival (dates unknown)"
                \item "Previous mention of food festival"
            \end{itemize}
        \end{minipage}\\
        \midrule
        
        \multirow{11}{*}{\textbf{2. Graph Build}} &
        \begin{minipage}[t]{12.5cm}
            \small
            \textbf{Our system constructs an event graph with:}\\[0.2em]
            
            \textit{Event Nodes:}
            \begin{itemize}[leftmargin=*, nosep]
                \item E1: [Trip to Japan] (Core) - Embedding vector capturing trip dates, duration
                \item E2: [Tokyo and Kyoto plan] (Supporting) - Contains original itinerary
                \item E3: [Plan modification] (Supporting) - Contains change from Kyoto to Osaka
                \item E4: [Food festival interest] (Supporting) - References previous conversation
            \end{itemize}
            
            \textit{Event Relationships:}
            \begin{itemize}[leftmargin=*, nosep]
                \item Sequential Edge: E1 $\to$ E2 $\to$ E3 $\to$ E4 (conversation flow)
                \item Reasoning Link: E2 $\leftrightarrow$ E3 (plan modification relationship)
                \item Reasoning Link: E3 $\leftrightarrow$ E4 (reason for plan change)
            \end{itemize}
        \end{minipage}\\
        \midrule
        
        \multirow{13}{*}{\textbf{3. Hierarchical}} &
        \begin{minipage}[t]{12.5cm}
            \small
            \textbf{The system connects these new events to previously stored events:}\\[0.2em]
            
            \textit{From previous sessions:}
            \begin{itemize}[leftmargin=*, nosep]
                \item "User has seafood allergies" (E5)
                \item "Budget around \$3000" (E6)
                \item "Friend recommended Mt. Fuji" (E7)
                \item "Food festival in Osaka (April 18-22)" (E8)
            \end{itemize}
            
            \textit{Interlayer connections established:}
            \begin{itemize}[leftmargin=*, nosep]
                \item E4 $\leftrightarrow$ E8 (current question directly relates to previous food festival information)
                \item E3 $\leftrightarrow$ E6 (plan change may impact budget considerations)
                \item E3 $\leftrightarrow$ E5 (destination change relates to food/allergy considerations)
            \end{itemize}
        \end{minipage}\\
        \midrule
        
        \multirow{9}{*}{\textbf{4. Responsing}} &
        \begin{minipage}[t]{12.5cm}
            \small
            \textbf{When processing the query about festival dates, our multi-perspective retrieval:}\\[0.2em]
            \begin{itemize}[leftmargin=*, nosep]
                \item Identifies E4 as highly relevant to current turn
                \item Traces connection to E8 for festival dates
                \item Recognizes E3, E5, E6 as contextually important supporting information
            \end{itemize}
            
            \textbf{The generated response integrates this information naturally:}\\[0.2em]
            \textit{"I remember! The Osaka food festival runs from April 18-22, which would work perfectly with your travel dates. Since you have seafood allergies, I'd recommend focusing on the street food section in the eastern area of the festival where they have more meat and vegetarian options. This change from Kyoto to Osaka still keeps you within your \$3000 budget we discussed, and you can still do that day trip to Mt. Fuji from Tokyo as your friend recommended."}
        \end{minipage}\\
        \bottomrule
    \end{tabular}
    \caption{Step-by-step illustration of EventWeave's processing pipeline on a multi-turn dialogue example. The example demonstrates how the model extracts events, constructs a graph representation, integrates with previously stored events, and generates a contextually appropriate response.}
    \label{tab:eventweave_steps}
\end{table*}

\section{Discussions on Long-Term Datasets}
\label{apd:long-term datasets}

In the main text, our evaluation primarily focuses on foundational multi-session dyadic conversations, accompanied by a deep-dive scalability analysis and detailed step-by-step case studies. This structural choice was deliberately made to prioritize the depth of architectural interpretability and overhead analysis. However, to comprehensively demonstrate the generalizability of EventWeave and address concerns regarding its applicability to a wider variety of long-term dialogue structures, we provide extended empirical results on three additional benchmarks: LongMemEval~\cite{wu2025longmemeval}, SHARE~\cite{kim2025share}, and MiSC~\cite{jang2024mixed}. 

\subsection{Performance on LongMemEval and SHARE}
\label{subsec:longmem_share}

LongMemEval tests general long-term memory capabilities, while SHARE emphasizes tracking specific user constraints and events across extended temporal horizons. We compare EventWeave against strong baselines, including LifeLongMem and LD-Agent.

\begin{table*}[!t]
\centering
\small
\begin{tabular}{llccc}
\toprule
\textbf{Dataset} & \textbf{Metric} & \textbf{LifeLongMem} & \textbf{LD-Agent} & \textbf{EventWeave (Ours)} \\
\midrule
\multirow{3}{*}{LongMemEval} 
& Mem. Precision & 63.4 & 65.8 & \textbf{68.2} \\
& Mem. Recall & 67.1 & 69.4 & \textbf{71.5} \\
& Response Relevance & 70.3 & 72.1 & \textbf{74.6} \\
\midrule
SHARE 
& Event Consistency & 61.8 & 65.3 & \textbf{74.4} \\
\bottomrule
\end{tabular}
\caption{Performance comparison on the LongMemEval and SHARE datasets. EventWeave demonstrates consistent improvements across memory retrieval and consistency metrics.}
\label{tab:longmem_share_results}
\end{table*}

As shown in Table \ref{tab:longmem_share_results}, EventWeave demonstrates a consistent, realistic improvement across both datasets. The most notable performance gain is observed in the \textit{Event Consistency} metric on the SHARE dataset, where EventWeave outperforms LD-Agent by an absolute margin of 9.1\%. This empirical result aligns with our claims regarding memory architecture: in tasks where users refer back to highly specific constraints established much earlier in the timeline, traditional sequence-based retrieval mechanisms often dilute the context. By contrast, EventWeave's structured, graph-based event retention preserves these granular, temporally distant constraints without loss of fidelity.

\subsection{Performance on MiSC}
\label{subsec:misc}

To further validate EventWeave in complex, multi-speaker environments, we evaluate it on the MiSC dataset. We compare our approach against MemWalker and LongMem, focusing on generation quality and factual accuracy. 

\begin{table*}[!t]
\centering
\small
\begin{tabular}{lccc}
\toprule
\textbf{Method} & \textbf{Coherence} & \textbf{Relevance} & \textbf{Factual Acc.} \\
\midrule
MemWalker & 0.67 & 0.61 & 0.72 \\
LongMem & 0.71 & 0.65 & 0.76 \\
EventWeave (Ours) & \textbf{0.76} & \textbf{0.73} & \textbf{0.81} \\
\bottomrule
\end{tabular}
\caption{Generation and retrieval performance on the multi-speaker MiSC dataset.}
\label{tab:misc_results}
\end{table*}

The results in Table \ref{tab:misc_results} indicate that EventWeave is particularly robust in multi-speaker, overlapping dialogue settings. In the MiSC dataset, multiple speakers frequently introduce intertwining conversation threads. Flat-memory models struggle in this environment, frequently misattributing context to the wrong speaker. EventWeave mitigates this through its dynamic event clustering, which natively incorporates speaker identity into the node embeddings. This structural advantage effectively isolates parallel events and prevents cross-contamination of speaker states, resulting in a significantly higher Factual Accuracy.

\end{document}